\documentclass[acmtog,screen]{acmart}

\usepackage{booktabs} 
\usepackage{wrapfig}
\usepackage{graphicx}
\usepackage{subcaption}
\usepackage{enumitem}
\usepackage{hyperref}
\setcopyright{none}

\usepackage[ruled]{algorithm2e} 

\SetAlFnt{\small}
\SetAlCapFnt{\small}
\SetAlCapNameFnt{\small}
\SetAlCapHSkip{0pt}

\copyrightyear{2025}
\acmYear{2025}
\setcctype{by-nc}
\acmConference[SIGGRAPH Conference Papers '25]{Special Interest Group on Computer Graphics and Interactive Techniques Conference Conference Papers }{August 10--14, 2025}{Vancouver, BC, Canada}
\acmBooktitle{Special Interest Group on Computer Graphics and Interactive Techniques Conference Conference Papers (SIGGRAPH Conference Papers '25), August 10--14, 2025, Vancouver, BC, Canada}
\acmDOI{10.1145/3721238.3730753}
\acmISBN{979-8-4007-1540-2/2025/08}

\begin{document}
\title{Photoreal Scene Reconstruction from an Egocentric Device}

\author{Zhaoyang Lv}
\email{zhaoyang@meta.com}
\orcid{0000-0002-7788-9982}
\affiliation{%
 \institution{Reality Labs Research, Meta}
 \country{United States of America}
}

\author{Maurizio Monge}
\email{maurimo@meta.com}
\orcid{0009-0000-0139-9146}
\affiliation{%
 \institution{Reality Labs Research, Meta}
 \city{London}
 \country{United Kingdom}
}

\author{Ka Chen}
\email{chenka@meta.com}
\orcid{0009-0009-9991-9757}
\affiliation{%
    \institution{Reality Labs Research, Meta}
    \city{Redmond}
    \country{United States of America}
}

\author{Yufeng Zhu}
\email{yufengzhu@meta.com}
\orcid{0009-0003-8716-201X}
\affiliation{%
 \institution{Reality Labs Research, Meta}
 \city{Redmond}
 \country{United States of America}
}

\author{Michael Goesele}
\email{research@goesele.org}
\orcid{0000-0002-0944-0980}
\affiliation{%
 \institution{Reality Labs Research, Meta}
 \city{Redmond}
 \country{United States of America}
}

\author{Jakob Engel}
\email{jakob.engel@meta.com}
\orcid{0000-0002-2799-5808}
\affiliation{%
 \institution{Reality Labs Research, Meta}
 \city{Redmond}
 \country{United States of America}
}

\author{Zhao Dong}
\email{zhaodong@meta.com}
\orcid{0000-0002-9026-6886}
\affiliation{%
 \institution{Reality Labs Research, Meta}
 \city{Redmond}
 \country{United States of America}
}

\author{Richard Newcombe}
\email{newcombe@fb.com}
\orcid{0009-0004-9091-8989}
\affiliation{%
 \institution{Reality Labs Research, Meta}
 \city{Redmond}
 \country{United States of America}
}

\renewcommand\shortauthors{Lv, Zhaoyang et al}

\begin{abstract}
In this paper, we investigate the challenges associated with using egocentric devices to photorealistic reconstruct the scene in high dynamic range. 
Existing methodologies typically assume using frame-rate 6DoF pose estimated from the device's visual-inertial odometry system, which may neglect crucial details necessary for pixel-accurate reconstruction. 
This study presents two significant findings. 
Firstly, in contrast to mainstream work treating RGB camera as global shutter frame-rate camera, we emphasize the importance of employing visual-inertial bundle adjustment (VIBA) to calibrate the precise timestamps and movement of the rolling shutter RGB sensing camera in a high frequency trajectory format, which ensures an accurate calibration of the physical properties of the rolling-shutter camera.
Secondly, we incorporate a physical image formation model based into Gaussian Splatting, which effectively addresses the sensor characteristics, including the rolling-shutter effect of RGB cameras and the dynamic ranges measured by sensors. 
Our proposed formulation is applicable to the widely-used variants of Gaussian Splats representation. 
We conduct a comprehensive evaluation of our pipeline using the open-source Project Aria device under diverse indoor and outdoor lighting conditions, and further validate it on a Meta Quest3 device. 
Across all experiments, we observe a consistent visual enhancement of +1 dB in PSNR by incorporating VIBA, with an additional +1 dB achieved through our proposed image formation model. Our complete implementation, evaluation datasets, and recording profile are available at \url{https://www.projectaria.com/photoreal-reconstruction/}
\end{abstract}


\begin{teaserfigure}
  \centering
  \captionsetup{type=figure}
    \includegraphics[width=1.0\linewidth]{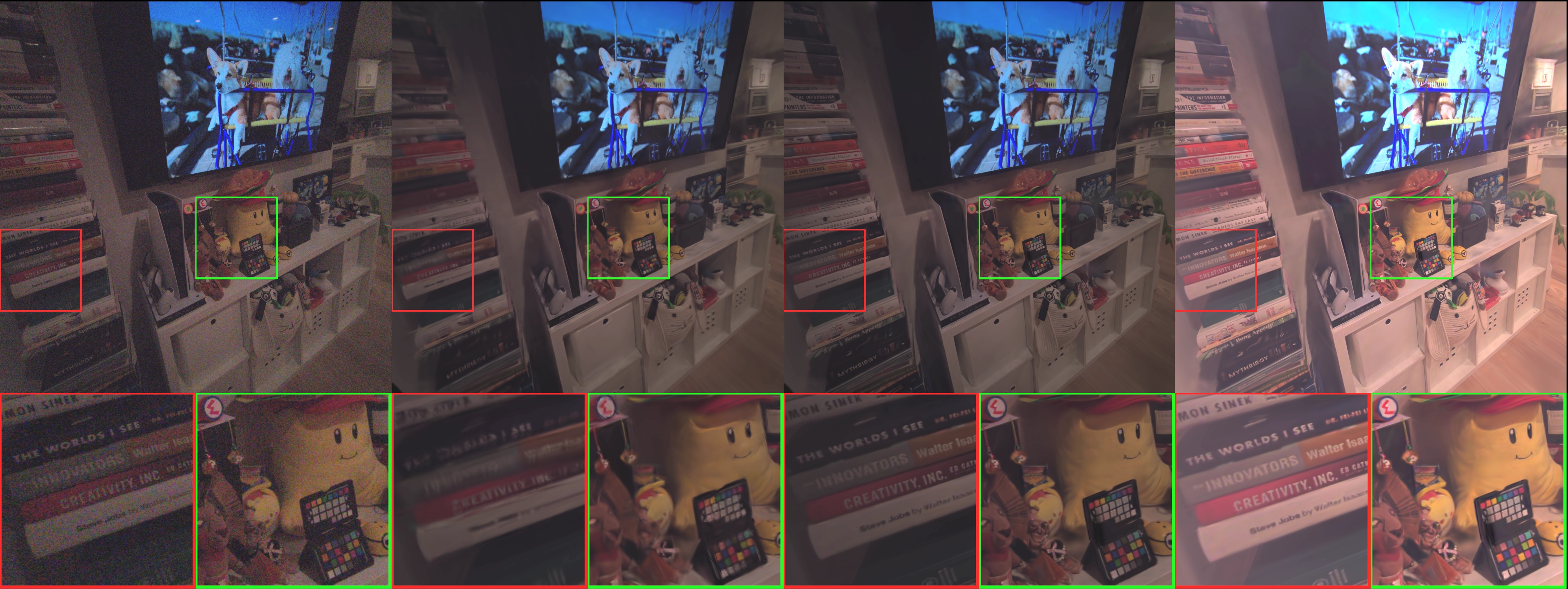}
   \caption{A comparison (from left to right) of (a) the raw camera held-out reference image, (b) the reconstructed view from a vanilla Gaussian-splatting algorithm using device calibration, (c) our reconstruction result using our proposed system, and (d) our reconstruction rendered with lifted dynamic range. We adjust the gamma of all comparisons for better visualization. Egocentric video may contain challenges in image quality due to high speed head motion and the form factor constraint. Our proposed system can recover photoreal scene reconstruction from the egocentric input videos with noises and heavy rolling-shutter effect. Note the improved clarity on the text in the low-light image areas due to our correct handling of rolling-shutter during both steps in visual inertial bundle adjustment and Gaussian-splatting. As a result, we can render the videos with higher dynamic range and further boost the details. }
   \label{fig:teaser}
\end{teaserfigure}


%
%


%
%


\maketitle


\newcommand{\scene}{\mathcal{S}}
\newcommand{\irradiance}{\mathbf{r}}
\newcommand{\pixel}{\textbf{u}}
\newcommand{\imageheight}{H}
\newcommand{\pose}{\textbf{T}}
\newcommand{\colorimage}{\textbf{C}}
\newcommand{\rasterizefunc}[1]{\pi(#1)}
\newcommand{\postprocessfunc}[1]{\phi(#1)}
\newcommand{\exposuretime}{t_e}
\newcommand{\readouttime}{\Delta d t_r}
\newcommand{\gain}{g}
\newcommand{\vignette}{V(\pixel)}
\newcommand{\rectifiedindex}{\mathcal{R}(\pixel)}
\newcommand{\gaussian}{\mathbf{G}}
\newcommand{\splatsposition}{\boldsymbol{\mu}}
\newcommand{\splatsrotation}{\mathbf{R}}
\newcommand{\splatsscale}{\mathbf{\mathbf{S}}}
\newcommand{\splatsopacity}{\alpha}
\newcommand{\splatscolor}{\mathbf{c}}
\newcommand{\covariance}{\boldsymbol{\Sigma}}
\newcommand{\real}[1]{\mathbb{R}^{#1}}
\newcommand{\SE}[1]{\text{SE}{#1}}
\newcommand{\SO}[1]{\text{SO}{#1}}




\section{Introduction}
\label{sec:intro}

Egocentric glasses equipped with first-person view cameras capture the world in an unparalleled perspective from the viewpoint of the human wearer.
Scalable photorealistic 3D reconstruction from these inputs holds significant potential for various applications in augmented reality, robotics, and spatial artificial intelligence.

Seminal work in neural rendering \cite{mildenhall2021nerf, muller2022instantngp, kerbl2023gaussiansplatting} has significantly advanced the field of photorealistic scene reconstruction using mobile cameras. Unlike traditional techniques that required depth sensing \cite{newcombe2011kinectfusion, straub2019replica}, neural rendering relies solely on posed images as input, making it well-suited to the form factor of egocentric glasses. Several studies have demonstrated that neural rendering facilitates 3D scene reconstruction and understanding \cite{gu2024egolifter, vadim2023epicfields}.

The constant motion of the human head presents significant challenges for image sensing and localization in egocentric devices, both of which can adversely affect the quality of neural reconstruction. 
Existing solutions employing visual-inertial odometry can provide a fast efficient solution for six degree of freedom (6DoF) tracking for the device, which is further used to estimate a frame-rate 6DoF pose for the RGB camera. However, a rolling-shutter camera is sensitive to the fast moving head and a frame-rate pose cannot accurately represent the correct pixel motions. 
Due to form factor constraints, the image quality captured by sensors lacking hardware stabilization can additionally be also compromised by potential motion blur and noise under undesired lighting conditions \cite{goesele2025plato}.

In this paper, we examine the challenges associated with egocentric sensing and propose a systematic 3D reconstruction framework for photorealistic novel-view rendering. Our approach includes specific details on device calibration, reconstruction methods, and the capture process. We utilize Project Aria \cite{engel2023aria}, an open-source egocentric glasses platform, for data capture, thereby providing representative data within appropriate form factor constraints. Furthermore, we demonstrate that insights gained from our study can be generalized to another commercial headset, such as the Meta Quest 3 device.

Our contributions are as follows:

Firstly, we address the importance of employing visual-inertial bundle adjustment (VIBA) that accounts for the rolling-shutter behavior of the RGB camera. This provides a continuous camera trajectory to model pixel movement in neural reconstruction.
Our experiments demonstrate that using VIBA will consistently improve the novel view quality in Gaussian Splatting by +1db in PSNR.

Secondly, we introduce a rasterization-based image formulation pipeline that addresses common artifacts in physical image formation, including rolling shutter, lens shading, exposure, and gain compensation. Our approach is distinct in that we represent image poses as posed pixel arrays sampled from a continuous trajectory, rather than assigning a single camera pose per image, and preserve the merit of Gaussian rasterization. 
Unlike existing methods that require ray-tracing Gaussians, e.g. \cite{nicolas20243dgrt}, our formulation is applicable to general-purpose rasterization-based Gaussian splatting. 
When being applied to 3D Gaussian Splatting (3DGS) \cite{kerbl2023gaussiansplatting}, our approach can further enhance reconstruction quality by +1 dB. We outperform existing baselines and demonstrate a large quality improvement in handling complex scenes observed by egocentric devices. 

Third, to reduce the effect of blur with rapid head motion in darker indoor scenes, we propose a strategy of deliberately underexposing input videos during the capture, inspired by HDR+ \cite{Hasinoff2016hdrplus}. 
We demonstrate we can reconstruct high quality noise-free scene radiance from noisy dim input videos, and further render sharp blur-free videos at a higher dynamic range.

We evaluate our algorithm on recorded datasets across scenes with different scale and complexity. In addition to existing public datasets, we recorded a new Aria scene dataset following our capture process to benchmark this study.
We release our code, dataset and capture profiles, with details at \href{https://www.projectaria.com/photoreal-reconstruction/}{Project Aria Photoreal Reconstruction}.



\section{Related Work}
\label{sec:related_work}

Since the seminal work NeRF \cite{mildenhall2021nerf} demonstrated novel view synthesis using radiance fields from posed images, numerous efforts have extended this to be faster \cite{muller2022instantngp,sara2022plenoxels}, with anti-aliasing \cite{barron2022mipnerf360}, at bigger scale \cite{barron2023zipnerf} or dynamic scenes \cite{li2022dynerf}.  
Different from a raytracing formulation in NeRF, Gaussian Splatting \cite{kerbl2023gaussiansplatting} introduced a new rasterization based neural radiance fields composed of anisotropic 3D Gaussians (3D-GS) efficiently reconstructs scenes at high quality and its extension show superior performance for geometry reconstruction as well \cite{huang20242dgs,yu2024gof}. Recent methods \cite{condor2024raytraced_primitives,nicolas20243dgrt,mai2024ever} use ray-tracing to optimize 3D-GS, addressing physical camera artifacts like rolling-shutter effects and deblurring, albeit with increased complexity. Our work addresses the physical camera properties using the rasterization based 3D-GS and shows it can improve the reconstruction quality on both 3D-GS and its variants.  

Image blur and noise are common in mobile camera captures and have been extensively studied in classical \cite{Hasinoff2016hdrplus} and learning based methods \cite{chen2018learningdark}. Multi-view images aid in understanding image formation from its principle, with seminal work in Richardson-Lucy deblurring \cite{liu2010richardsonlucy} and HDR+ \cite{Hasinoff2016hdrplus}. 
This concept extends to neural radiance fields. RawNeRF \cite{mildenhall2022rawnerf} reconstructed radiance fields from noisy low dynamic range input and outperformed single and multi-image raw denoisers. Approaches have explored reconstruction from blurred inputs using NeRF \cite{ma2021deblurnerf, wang2023badnerf} or Gaussian splatting \cite{zhao2024badgaussians,lee2024deblurring3dgs,seiskari2024gaussian_on_the_move}, with a similar concept applicable to rolling-shutter as well \cite{niu2024rsnerf}. They estimate the scene radiance with camera ray deformation induced by motion. However, due to unknown sensor calibration for exposure or rolling-shutter, they require jointly optimizing camera deformation along with the scene radiance. One recent approach \cite{seiskari2024gaussian_on_the_move} levarages high frequency visual inertial tracking and proposed to rasterize with additional pixel velocity measured against visual aligned IMU information.  

Our approach is most related to the synthesized capabilities from RawNeRF \cite{mildenhall2022rawnerf} and 3DGS-on-move \cite{seiskari2024gaussian_on_the_move}. Unlike RawNeRF, we used Gaussian Splatting for the scene representation and addressed the rolling-shutter in egocentric video beyond camera noise. 
Our solution is based on high-frequency tracking for the camera sensors. It addresses the pose acquisition issue to apply RawNeRF or its variants \cite{singh2024hdrsplat} in practice at scale. 
Our method leverages high-frequency tracking from VIO, explicitly modeling rolling-shutter and blur as in 3DGS-on-move \cite{seiskari2024gaussian_on_the_move}. Unlike 3DGS-on-move, which requires rasterizing pixel velocities with pinhole assumption for rolling-shutter and jointly pose optimization during training, our approach applies to any Gaussian Splatting variant with lens distortion, and we demonstrate superior quality consistently across scenes.

Neural rendering methods are commonly evaluated using static multi-view images captured by devices that uses image signal processing (ISP) \cite{mildenhall2019llff,barron2022mipnerf360,barron2023zipnerf}, which includes non-transparent processes of denoising, deblurring, sharpening and tone mapping that cannot be inverted. The camera calibration and poses are acquired using off-the-shelf structure from motion tools, such as COLMAP \cite{schoenberger2016colmap}. For images collected with extreme blur \cite{ma2021deblurnerf} or noise, the success in acquiring such camera ground truth can be extremely volatile. For successfully localized scenes, this calibration process also lacks physically meaningful calibration information to study the impact of 3D motions. 
3DGS-on-the-move \cite{seiskari2024gaussian_on_the_move} collected hand-held videos with synchronized IMUs and demonstrated the first 3D-GS to jointly optimize rolling-shutter and motion-blur aware poses using imperfect factory calibration as initialization. 
Compared to all existing work, we propose to address these challenges in an egocentric device systematically from the input capture procedure to reconstruction algorithm. 
We evaluate the system robustly across different scenarios and our data acquisition can be reliably reproduced using open-source egocentric device platform. 

In egocentric vision, neural scene reconstruction has also served as an important building block to enable scene understanding \cite{straub2024efm3d, gu2024egolifter, vadim2023epicfields} and contextual AI with human interaction \cite{lv2024aea,yi2024egoallo,plizzari2024spatialcognition}. However, it has been a persistent challenge to acquire high quality reconstructions from egocentric devices in these prior works using existing algorithms. We believe that our work can pave the way for a scalable and accessible high quality neural reconstruction in the rapid growing egocentric research areas.



\section{Background}
\label{sec:background}


\begin{figure}[t]
\centering
\includegraphics[width=0.5\textwidth]{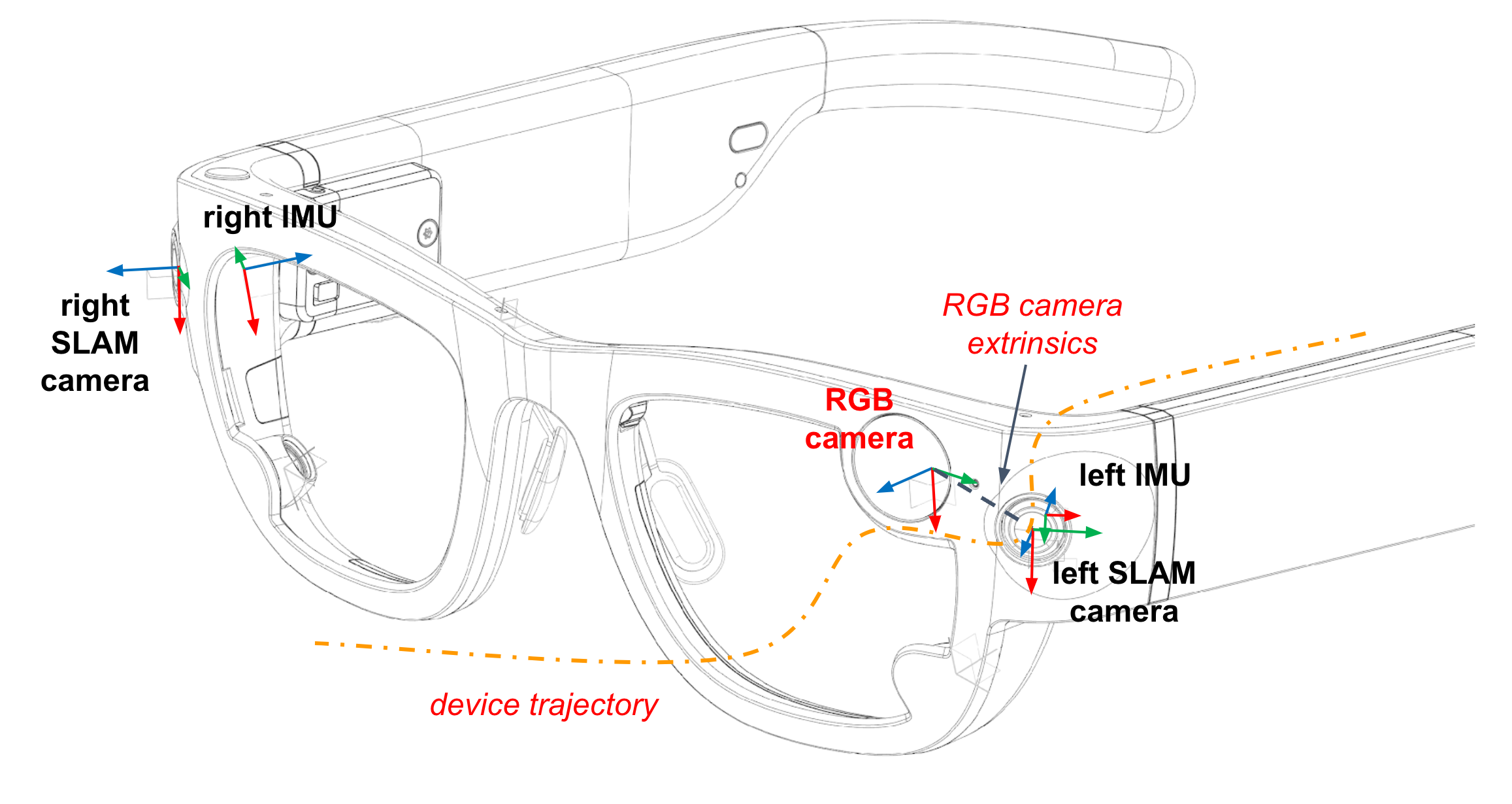}
\caption{The layout of sensors for state estimation in a Project Aria device. The device trajectory is represented at high frequency at IMU rate. The red color highlighted the input information used within Gaussian Splatting reconstruction.}
\label{fig:sensors}
\end{figure}


One of the key distinguishing characteristic we present in this paper is to \textbf{focus on reconstructing a video captured by an egocentric form factor device}. This is different from commonly used reconstruction datasets composed of static image snapshots using phones \cite{mildenhall2019llff, barron2022mipnerf360}, professional cameras \cite{barron2023zipnerf} or high-end 3D scanners \cite{knapitsch2017tandt}. 

We use the open platform Project Aria \cite{engel2023aria}, which is a representative egocentric device with form factor and software for future commodity 3D sensors. We consider the following properties as the essential inputs of our study: 

\setlength{\topsep}{0pt}
\begin{enumerate}[leftmargin=*]
\setlength\itemsep{0em}    
    \item With a high-frequency closed-loop device trajectory at the IMU rate (e.g., 1 kHz for Project Aria), we can approximate the 6DoF poses as a piecewise continuous function with respect to timestamps. The raw sensor measurements are timestamped on a common clock at nanosecond resolution. 
    With the state estimation as described in Section~\ref{sec:state_estimation}, we can reliably calculate the asynchronous posed rays using the estimated pixel timestamp derived from first principles.    
    \item The device provides a raw sensor output for the RGB camera, including parameters such as gain, exposure, and a calibrated vignette image. This facilitates the modeling the physical image formation process without approximation. Our captured images do not undergo any additional image signal processing (ISP), such as denoising or local tonemapping.
\end{enumerate}

In the following, we will discuss the importance of acquiring high-frequency RGB sensor calibration in Section~\ref{sec:state_estimation}. Then we will introduce a Gaussian Splatting pipeline that leverages the high frequency trajectory and raw sensor models in Section~\ref{sec:method}. We will further discuss details in capture settings that can improve scene reconstruction in Section~\ref{sec:capture_data}.



\section{Visual Inertial Bundle Adjustment}
\label{sec:state_estimation}


\begin{figure}[t]
\centering
\includegraphics[width=0.5\textwidth]{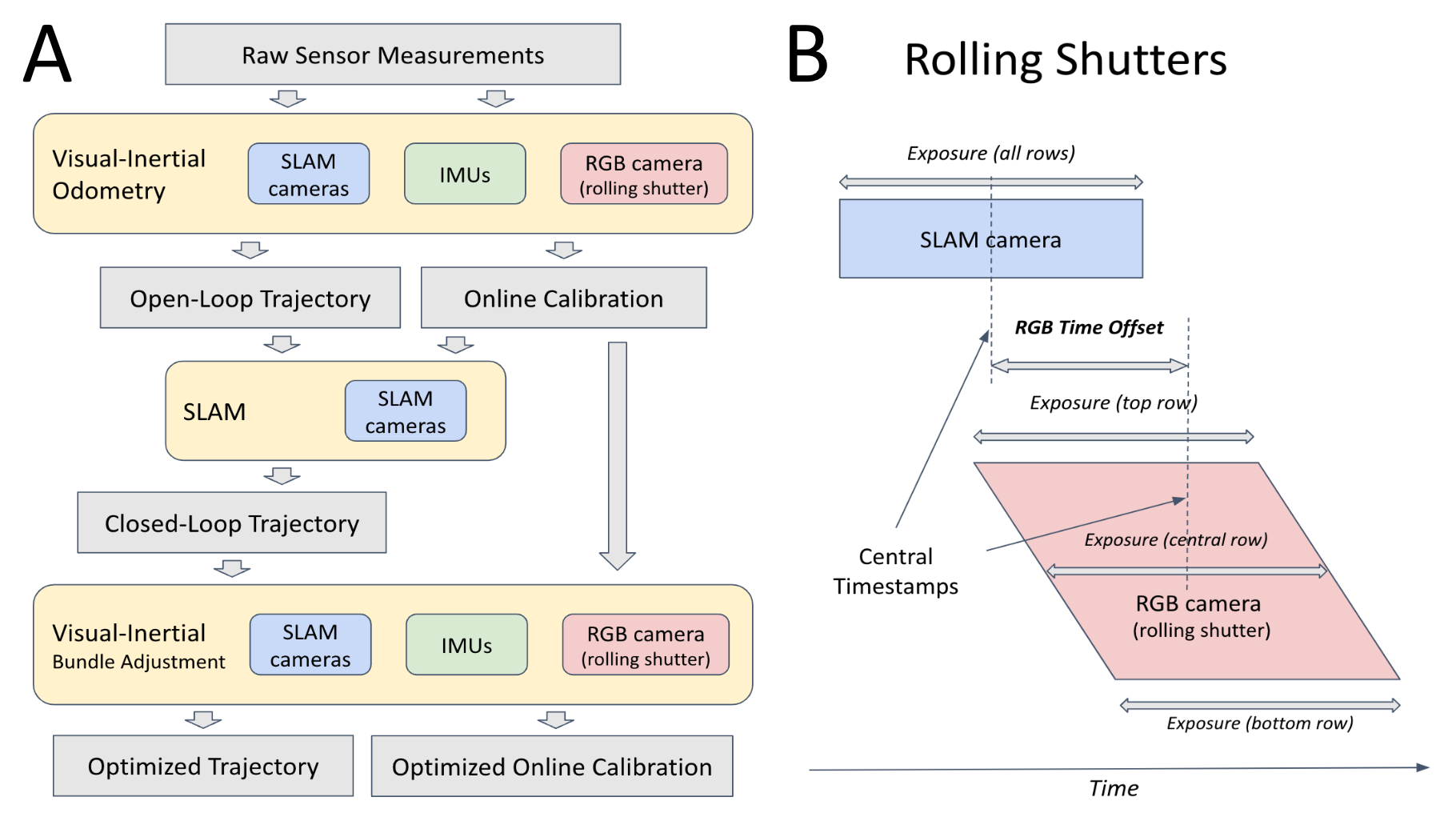}
\caption{\textbf{A.} An overview of the state estimation pipeline. Among them, the VIBA process handle the rolling-shutter RGB camera in a global bundle adjustment. We provide an exemplification of the rolling shutter properties in \textbf{B.}, which are handled in the VIBA step. VIBA models the rolling-shutter RGB camera and outputs accurate timestamps with poses for pixel exposed at different rows during the readout time.}
\label{fig:pipeline}
\end{figure}


The state estimation pipeline of an egocentric device (Fig. \ref{fig:pipeline}) contains high-frequency (1KHz) device trajectory and online sensor calibration values. 
In Project Aria device, the sensors (Fig. \ref{fig:sensors}) used for estimation are, additionally to the RGB camera, two monochrome global-shutter SLAM cameras and two inertial motion units (IMU).
The full state estimation in our system follows the following steps:

\begin{enumerate}[leftmargin=*]
    \item The visual inertial odometry (VIO) system fuses sensor measurements computing an incrementally estimated (open-loop) frame-rate device trajectory and online calibration of all sensors.
    \item The SLAM system uses monochrome global-shutter SLAM cameras to provide loop closures and multi-recording relocalizations, computing a closed-loop trajectory.
    \item Resulting estimate is batch-optimized in a \textbf{visual-inertial bundle adjustment (VIBA)} step, which improves the accuracy while maintaining loop-closing constraints. Trajectories formed by all frames, including RGB frames and all sensor calibrations are re-estimated in a joint optimization step. 
\end{enumerate}

Step (3) provides the essential calibration for the RGB cameras that our reconstruction system depends on, which we will refer to as \emph{VIBA}. Compared to an alternative off-the-shelf bundle adjustment system such as COLMAP \cite{schoenberger2016colmap}, VIBA has a few essential differences:

\begin{enumerate}[leftmargin=*]
  \item RGB camera calibration uses a rolling-shutter aware model jointly with global-shutter SLAM cameras.
  \item Camera time offsets are optimized as part of the model when the hardware is unable to provide accurate trigger times.
  \item VIBA re-estimates all calibration parameters at IMU frequency for improved estimate accuracy while simultaneously making reprojection errors sub-pixel on tracked points.
  \item The system scales well to long egocentric videos. 
\end{enumerate}

The output of VIBA ensures a high-frequency rolling-shutter aware image formation model, which existing system does not provide. To the best of our knowledge, no existing work investigated the importance of this feature for 3D reconstruction, and the improvement of calibration estimate is especially relevant in the regions that are poor of tracked points, where mismatch might occur if we only tried to improve reprojection errors on the sparse set of tracked points. 
The SLAM pipeline, including the step with VIBA, is now accessible through the machine perception service in Project Aria tools. The Project Aria Docs website offers comprehensive instructions on how to access and utilize this tool. We employed the publicly available tool to obtain all input in the paper.



\section{Methods}
\label{sec:method}

We use Gaussian Splatting (3D-GS) \cite{kerbl2023gaussiansplatting} as the scene representation, a popular framework for efficient photorealistic scene reconstruction. Unlike existing approaches \cite{seiskari2024gaussian_on_the_move,nicolas20243dgrt}, our proposed approach handles camera motions such as rolling shutter, and lens distortion with no change required in the standard Gaussian rasterization and can be applicable to its broad family of advanced variants. 

We first discuss a few key notations in 3D-GS and its variants. Then we will illustrate how we update its image rasterization formulation model to handle the common artifacts in egocentric camera sensing. We provide the full implementation details of the preprocessing steps of all input data in the supplementary material. 

\subsection{Gaussian Splatting}

The 3D-GS represents the scene $\scene$ as a set of 3D Gaussians 
$\gaussian = \{\splatsposition, \splatsrotation, \splatsopacity, \splatscolor\}$. 
Each Gaussian is determined by its 3D mean position $\splatsposition \in \real{3}$ and 3D covariance $\covariance \in \real{3\times3}$. To approximate a semi-definite 3D covariance, it is parameterized as $\splatsrotation \splatsscale \splatsscale^{T} \splatsrotation^{T}$ using rotation $\splatsrotation \in \SO{3}$ and scale $\splatsscale \in \real{3}$. To render an image, all 3D Gaussians are first projected to 2D given the camera pose $\pose \in \SE{3}$ as sorted 2D Guassians according to the projected depth value. The pixel color $\colorimage(\pixel)$ is an accumulation of the Gaussian color $\splatscolor \in \real{3}$ and opacity value $\splatsopacity \in \real{1}$ by traversing the list front-to-back. To simplify the notation, we represent the rasterization process to acquire the color of a pixel using the following rendering function: 
\begin{equation}
\label{eq:gaussian_rasterization}
    \colorimage(\pixel) = \rasterizefunc{\pixel, \scene, \pose}
\end{equation}
The above rasterization function in Eq.~\ref{eq:gaussian_rasterization} can also generalize to other Gaussian alternative, e.g. 2D-GS \cite{huang20242dgs}, with slight different parameterization of the scene $\scene$. We will use this function in Eq.~\ref{eq:gaussian_rasterization} to refer the broad family of Gaussian rasterization approaches in following sections and results. 

\subsection{Image rasterization model with high frequency poses}
\label{sec:physical_image_rasterization}

We represent the high frequency trajectory as a piecewise continuous function with the 6DoF pose, which support the pose query at any time $t$ as $\pose(t) =f_{\pose}(t)$. For a rolling-shutter camera, each row of a pixel has its asynchronous 6DoF pose given the query time at $t(\pixel)$ from from the image capture time $t(\textbf{0})$ and readout time $\readouttime$:
\begin{equation}
\label{eq:rolling_shutter_query_time}
    t(\pixel) = t(\textbf{0}) + (\frac{\pixel_h}{\imageheight}) \cdot \readouttime
\end{equation}
where $t(\textbf{0})$ represent the capture time of the first row pixel for the image with height $\imageheight$ and $\pixel_h$ is its row index.  

The physical image formation for each pixel accumulates photons of projected scene irradiance during a fixed exposure time $\exposuretime$ and amplified by an analog or digital gain value $\gain$. The image irradiance is further transformed and compressed into an image with certain dynamic range. We can calculate the color of each pixel $\colorimage(\pixel)$ by all the sampled pixels with poses $\pose$ starting $t(\pixel)$ within the sampled exposure interval $\exposuretime$, as the following updated function: 
\begin{equation}
\label{eq:color_rasterization}
   \colorimage(\pixel) = \postprocessfunc{\omega(\pixel)\int_{0}^{\exposuretime} \rasterizefunc{ \pixel, \scene, \pose(t(\pixel)+t)} dt} 
\end{equation}
where $\omega(\pixel)$ is a per-pixel linear weight that combines the effect of analog gain $\gain$, lens shading $\vignette$ and normalization factor. $\postprocessfunc{\cdot}$ represents the camera response function. We will discuss a few importance factors in practice as following.

\paragraph{Rolling-shutter with lens distortion:} The query time in Eq.~\ref{eq:rolling_shutter_query_time} is calculated based on the raw undistorted image. After image rectifiction, the linear relationship for each pixel respect to their row number will not hold which prohibits existing solution \cite{seiskari2024gaussian_on_the_move} to be applicable. We propose to generate a index ratio as a look-up table for the rectified image $\rectifiedindex$, where each pixel value is a ratio representing its row number relative to the image height. We adjust Eq.~\ref{eq:rolling_shutter_query_time} to the following time query:
\begin{equation}
\label{eq:general_rolling_shutter_query_time}
t(\pixel) = t(\textbf{0}) + \rectifiedindex \readouttime
\end{equation}

\paragraph{Camera motion sampling:} A discrete form of Eq.~\ref{eq:color_rasterization} requires sufficiently sampling the possible motions that could result in substantial pixel offsets during the exposure or readout time. 
We use the scene depth to estimate the reprojection error during a temporal bracket. Specifically, the sparse depth value is triangulated from the tracked points in global shutter SLAM cameras and serve as the 3D anchors for camera motion estimation. We estimate the length of temporal bracket that ensures half of the reprojected pixels have fewer than 1 pixel reprojection error. As a result, the full-resolution RGB camera with a readout time of approximately 16 ms has an average of eight motion samples within the readout time.


\begin{figure}[t]
    \centering
    \includegraphics[width=0.49\textwidth]{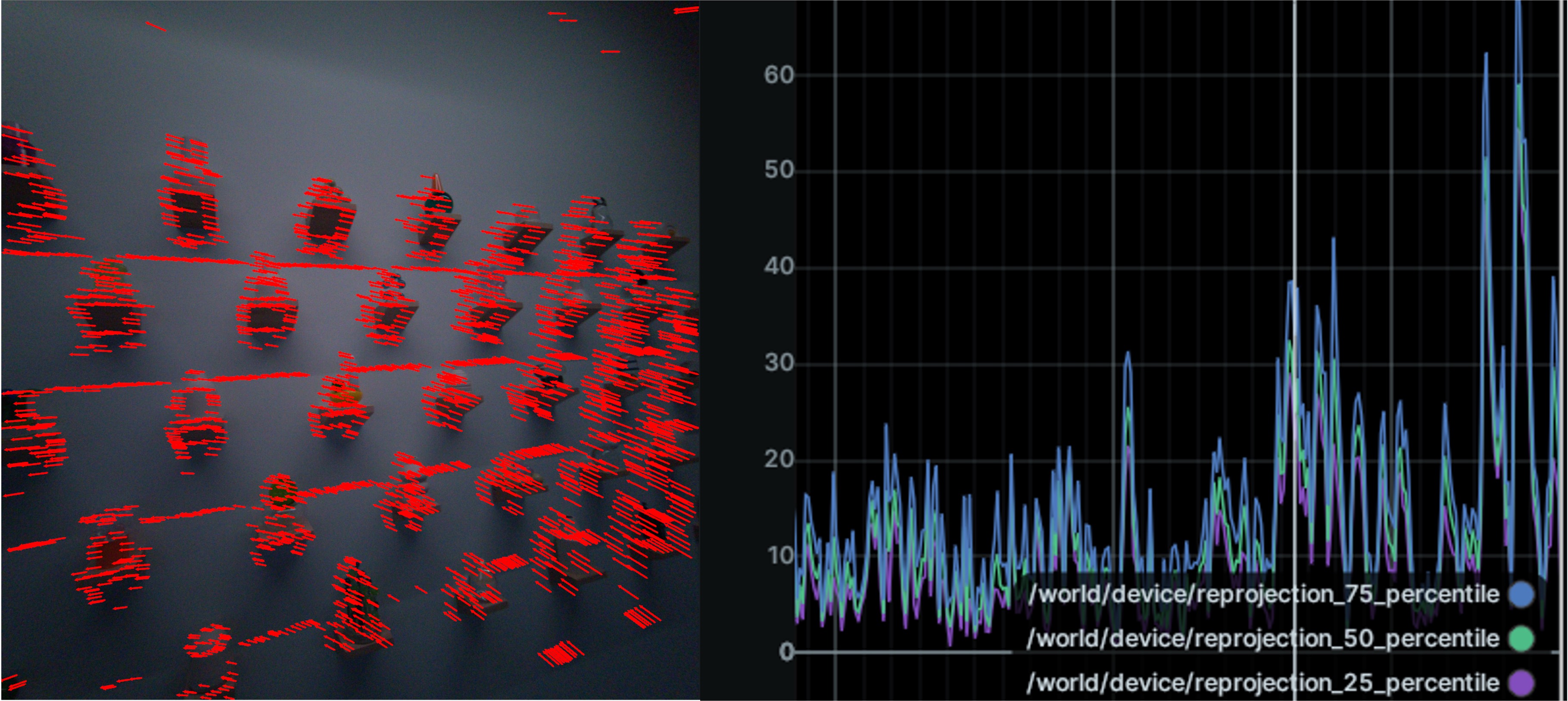}
    \caption{We visualize the impact of motion during image read-out time. In the left image, we visualize the reprojection vector of sparse scene depth during the image read out time. On the right, we calculated the magnitude of reprojection error for all points per-frame, and plot the 25,50,75 percentile of distribution along this trajectory in time. In this particular frame, 50 percentile of the points have about reprojection error of 30. Such errors will cause misalignment in reconstruction if not handled properly.}
    \label{fig:camera_motion_artifacts}
\end{figure}


\paragraph{The importance of motion sampling:} The human head is in constant motion and can achieve rotational velocities of several hundred degrees per second. Fig.~\ref{fig:camera_motion_artifacts} shows the artifact of pixel reprojection errors during the image readout time. In our ablation study, we will show correcting modeling the pixel motions can drastically improve the reconstruction quality.

\paragraph{Rasterization:} We batch rasterize the image based on the number of samples in $\rectifiedindex$ during the forward process and then use a gather operation to synthesize a final image. The number of pixels contribute to the backward gradients are same as the single image. We employ the aforementioned camera motion sampling strategy to determine the sample bracket within $\rectifiedindex$. For a quasi-static viewpoint, only one pose sample is needed, whereas 8-16 samples may be required for a fast-moving view.  

This approach is general to camera models and can be particular helpful for those containing high-order distortions. In our example, the RGB camera is a fisheye model with high order of coefficients which no existing rasterization implementation supports. 
Unlike existing approach that require customized rasterization kernel \cite{seiskari2024gaussian_on_the_move} to camera with particular calibration or using raytracing  \cite{condor2024raytraced_primitives}, our proposed rasterization in Eq.~\ref{eq:color_rasterization} with rectified index look up table require no change to general 3D-GS or its variants such as 2D-GS \cite{huang20242dgs} using common pinhole or fisheye rectified images \cite{liao2024fisheyegs}. 

\subsection{Additional factors}
\label{subsec:additional_factors}

\paragraph{Preserving dynamic range of the scene:} Different from datasets that often estimate the scene radiance directly using tonemapped images, the Gaussian color $\splatscolor$ of the scene $\scene$ in Eq.~\ref{eq:color_rasterization} is in linear space by default, which can hurt the optimization particular in high dynamic range scenarios.
Thus, we explicitly encode scene color as a gamma compressed scene irradiance $\splatscolor = \phi^{-1}(\irradiance)$. We use the same notation $\postprocessfunc{\cdot}$ to represent the inverse of camera response for simplicity. We use gamma value 2.2 as our default setting. 

\paragraph{Image blur:} For an auto-exposed camera, the exposure time is related to the auto-exposure target and the scene luminance. In bright outdoor scenarios, where the exposure of 0.5 ms is sufficient, the amount of motion during the exposure time is limited. However, in indoor scenes, the exposure can be ten times longer, leading to significant motion blur. Although Eq.~\ref{eq:color_rasterization} models the blur formulation and can be used to optimize deblurred pixels as existing work \cite{seiskari2024gaussian_on_the_move}, we find that it is challenging to reliably reconstruct sharp details from the blurry image without a very dense capture, which is hard to be fulfilled by human movement in practice. Instead, we can record less blurry images with shorter exposure time, which motivated our capture process to be discussed in Section \ref{sec:capture_data}.

\paragraph{Handling noise:} The Project Aria device uses analog gain to compensate image brightness in low light scenario resulting in strong photon shot noise that affects the image quality. For a gamma compressed image, an approximated square-root gamma value can whiten this photon shot noise and can be handled by the L1/L2 reconstruction loss without additional efforts \cite{lehtinen2018noise2noise}. Different from existing HDR reconstruction \cite{mildenhall2022rawnerf,singh2024hdrsplat} that target a HDR recovery using special losses for extreme dark scenario, we found the standard 3D-GS training objective is effective handle such indoor scenarios.



\section{Capturing Egocentric Video}
\label{sec:capture_data}

We record egocentric videos as 10 FPS JPEG-compressed 8MP images using Project Aria \cite{engel2023aria}. No denoising or deblurring is applied to the input video. We rectified the images to 2400x2400 resolution and use them for training.

In indoor scenario with illumination under 300 lux, standard auto-exposure system will increasing either the exposure time or the analog gain. Increasing the exposure time can result in motion-blurred captures, while increasing the gain can introduce more noises. In Section \ref{subsec:additional_factors}, we discuss certain level of photon shot noise can be handled via a noise-to-noise reconstruction process while it remains challenging to handle blur.
This inspired us to use a special capture treatment to handle low light scenarios. 

\paragraph{Capturing videos in indoor environment.} Inspired by high dynamic range (HDR) image processing algorithms that recover HDR images from multiple fast exposure captures \cite{Hasinoff2016hdrplus}, we propose capturing videos using fast exposures to minimize the impact of motion blur and recover high dynamic range, noise-free 3D scenes from the 3D reconstruction. We limit the RGB camera exposure time to a maximum of 2 ms. This works for general indoor environments measured with illuminations between 150-300 lux. Low light indoor scenarios is extremely challenging for egocentric video which is a limitation we leave for future work.

\paragraph{Aria scene evaluation dataset.} We collected two categories of egocentric recordings to evaluate the algorithm in diverse conditions. Each has 6 recordings in scenes with varying complexity. The \emph{outdoor recordings} have high dynamic range illuminated with a minimum of 3K lux. We used auto-exposure with varying exposures and minimum analog gain. Little motion blur or noise is present in these recordings. We use them to evaluate free viewpoint reconstruction in unconstrained large-scale environments. For \emph{indoors recordings}, we collect them with the proposed indoor capture protocol. Fig.~\ref{fig:qualitative_comparisons_baselines} and Fig.~\ref{fig:qualitative_comparisions_ablation} contain some visual examples and evaluation. We include more details of the dataset in supplementary materials.

The full processing of the dataset is based on tools from the open-sourced Project Aria platform.
We release the collected datasets as examples and we hope it can help the community can bring the learnings to various applications at a bigger scale.




\begin{table*}[h]
    \centering
    \caption{Quantitative evaluations for the Aria scene dataset. We separate the scenarios for outdoor and indoor scenarios. Splatfacto refers to \cite{ye2024gsplats} and 3DGS-on-move refers to \cite{seiskari2024gaussian_on_the_move}.}
    \begin{tabular}{l|cc|cc|cc|cc|cc|cc}
    \toprule
     \textbf{Ourdoor scenes}  & \multicolumn{2}{c}{bike shop}   & \multicolumn{2}{c}{steakhouse patio} & \multicolumn{2}{c}{pop-up shop} & \multicolumn{2}{c}{sunroom}     & \multicolumn{2}{c}{garden} & \multicolumn{2}{c}{restaurant patio} \\
       & PSNR           & SSIM           & PSNR              & SSIM              & PSNR           & SSIM           & PSNR           & SSIM           & PSNR           & SSIM           & PSNR               & SSIM              \\ 
    \midrule
    splatfacto                     & 26.98 &  0.803 &  26.30 &  0.777  & 26.48 &  0.766 & 22.82 & 0.735 & 20.32 & 0.647 & 20.34 & 0.701 \\
    3DGS-on-move  & 27.07 & 0.806  & 25.80 &  0.774  & 26.91 &  0.773 & 23.33 & 0.752 & 19.68 & 0.644 & 20.33 & 0.706  \\    
    \midrule
    Our Egocentric-GS                                  & \textbf{29.98} & \textbf{0.838} & \textbf{28.38}  & \textbf{0.805}    & \textbf{29.03} & \textbf{0.797}   & \textbf{27.03} & \textbf{0.788} & \textbf{24.00} & \textbf{0.704}  & \textbf{25.30}     & \textbf{0.787}    \\
    w/o VIBA                                            & 27.68         & 0.800           &   27.03  &  0.782                   &  27.64        &  0.771            &   25.22         &    0.751        &   22.58      &  0.670               &    22.32   &     0.725    \\
    w/o motion sampling                                  & 28.83          & 0.819          & 27.70    & 0.792                    & 28.29         & 0.783             & 26.15          & 0.766          &   22.90      &   0.678          &    23.31     &      0.746           \\
    w/o scene gamma                                  & 29.04          & 0.836          & 27.21             & 0.790             & 28.16          & 0.795          & 21.76          & 0.746          & 21.74          & 0.685          & 22.52              & 0.765             \\
       \midrule
       \midrule 
     \textbf{Indoor scenes}  & \multicolumn{2}{c}{Library}   & \multicolumn{2}{c}{Plant hallway} & \multicolumn{2}{c}{Open hallway} & \multicolumn{2}{c}{Micro Kitchen}     & \multicolumn{2}{c}{Multi-Floor} & \multicolumn{2}{c}{Livingroom} \\
       & PSNR           & SSIM           & PSNR              & SSIM              & PSNR           & SSIM           & PSNR           & SSIM           & PSNR           & SSIM           & PSNR               & SSIM              \\  \midrule  
    splatfacto                    & 23.62 & 0.522                     & 21.65 & 0.723                     &  23.85 & 0.530                            &  23.05 & 0.547                    & 24.92 & 0.520                     & 27.41 & \textbf{0.546}  \\
    3DGS-on-move & 22.03 & 0.501                     & 20.449 & 0.712                    &  21.85 & 0.511                            &  23.35 & 0.544                    & 23.99 & 0.511                     & 27.62 & \textbf{0.546} \\
    \midrule
    Our Egocentric-GS                                  & \textbf{24.59} & \textbf{0.544}   & \textbf{23.97} & \textbf{0.765}   &  \textbf{26.61} & \textbf{0.554}   &  \textbf{27.11} & \textbf{0.579}  & \textbf{25.96} & \textbf{0.535}  & \textbf{27.73} & \textbf{0.546}     \\
    w/o VIBA                                              & 23.91 & 0.525                     & 22.64 & 0.731                     & 25.63 & 0.539                             &  25.54 & 0.559  & 25.26 & 0.522 & 26.35 & 0.527  \\
    w/o motion sampling                                  & 23.98 & 0.525                     & 22.82 & 0.740                     & 25.66 & 0.540                             &  25.94 & 0.565 & 25.57 & 0.527                              & 27.28 & 0.535      \\
    w/o scene gamma                                      & 24.22 & 0.542                     & 21.02 & 0.744                     & 25.01 & 0.547                             & 24.48 & 0.561                    & 25.66 & \textbf{0.535}     & 27.33 & 0.545                     
    \\ \bottomrule
    \end{tabular}
    \label{tab:aria_scene}
    \vspace{1.5em}
\end{table*}


\section{Experiments}
\label{sec:experiments}

\paragraph{Baselines and ablations.} We use 3D-GS in GSplats \cite{ye2024gsplats} as our main rasterization method to represent the family of Gaussian Splatting algorithms. For all comparisons, we use the same hyperparameters following the baseline. We also include results that use 2D-GS as the rasterization method, which demonstrates our method is applicable to other Gaussian variants. We provide the following evaluation. 
\begin{enumerate}[leftmargin=*]
    \setlength\itemsep{0em} 
    \item \textbf{Splatfacto.} The Splatfacto in Nerfstudio integrates a few advanced features of Gaussian Splats. We use the same GSplats version (1.4) in our training and this baseline for a fair comparison. We provide the same calibration for RGB camera that for our ablations that does not use VIBA, which represents the most common reconstruction baseline in existing work \cite{gu2024egolifter, yi2024egoallo} using egocentric devices.
    \item \textbf{3DGS-on-move \cite{seiskari2024gaussian_on_the_move}.} This represents the state-of-the-art work that similarly rasterizes the Gaussians with an explicit image formation model. It rasterizes the Gaussians from a moving camera using the camera velocity information. In addition, it jointly optimizes the camera parameters during training to correct potential pose errors. We calculated the RGB camera velocity and provide them to this algorithm as initialization. Other than this, it use the same input as Splatfacto. We use the default parameters setting and the same training iterations as all other baselines. 
    \item \textbf{Ours Egocentric GS (3D-GS/2D-GS).} This is our implementation using the extensions we discussed in Section~\ref{sec:physical_image_rasterization}, based on closed-loop device trajectories and online calibrations. We use 3D-GS as the default choice of Gaussian rasterization pipeline if not clarified. 
    \item \textbf{Ablations: without VIBA.} We use the high frequency closed-loop trajectory calculated using the SLAM cameras and the factory calibration for RGB camera before VIBA. As splatfacto, it represents the commonly used setting in previous work.
    \item \textbf{Ablations: without motion sampling.} We disable motion aware pose sampling technique in Section \ref{sec:method} to compensate the rolling shutter effect. As all other baselines, we only use the center row pose represent the image pose, calculated from the VIBA calibrated trajectory. 
    \item \textbf{Ablations: without scene gamma.} In Eq.~\ref{eq:color_rasterization}, we represent scene radiance in linear space with no gamma conversion. 
\end{enumerate}

\paragraph{Evaluation datasets.} We evaluate our algorithm using the following data:
\begin{enumerate}[leftmargin=*]
    \setlength \itemsep{0em}
    \item \textbf{Aria scene dataset.} To evaluate the algorithm performing at scale, we collected a set of indoor and outdoor egocentric videos using the protocols discussed in Section~\ref{sec:capture_data}.  
    \item \textbf{Digital Twin Catelog (DTC) dataset.} We use the egocentric recordings within the Digital Twin Catalog dataset \cite{dong2025dtc}, which has precisely 3D aligned ground truth. These videos are recorded with the same lab lighting condition and fixed exposure gain inputs using Project Aria device. We evaluate the predicted geometry quality using the depth and normal rendered from the 3D ground truth.
    \item \textbf{Quest scene dataset.} We further collect one sequence using the Meta Quest 3 device. We process the dataset following the same proposed process. This evaluation demonstrate the generalization of the proposed method to cope with other data recorded from other commodity egocentric headsets.
\end{enumerate}

For all of the evaluations, we create the validation set following the common practices as \cite{mildenhall2019llff} that held out every 8th image as the validation images and use the rest for training. We perform all evaluations using PSNR and SSIM as the main image quality metrics.


\begin{table}[t]
\centering
\small
\caption{Quantitative evaluation of appearance and geometry reconstruction on egocentric DTC dataset. \cite{dong2025dtc}. We evaluate depth using scale-invariant L1 loss and evaluate normal using L1 loss.}
\begin{tabular}{lccc}
\toprule
      & PSNR $\uparrow$ & Depth $\downarrow$  & Normal $\downarrow$  \\ \midrule
Ours (3D-GS) & \textbf{29.83} & \textbf{0.1505} & \textbf{0.3078} \\
w/o motion sampling & 28.93 & 0.1769 & 0.3171 \\
w/o VIBA            & 28.52 & 0.1730 & 0.3274 \\ 
w/o scene gamma     & 28.76 & 0.1768 & 0.3236 \\
splatfacto \cite{ye2024gsplats} & 24.81 & 0.1749 & 0.6627 \\
\midrule 
\midrule
Ours 2D-GS & \textbf{29.54} & \textbf{0.1474} & \textbf{0.1509} \\
w/o motion sampling & 28.75 & 0.1755 & 0.2112 \\ 
\bottomrule
\end{tabular}
\label{tab:dtc_evaluation}
\end{table}

\begin{figure*}
    \centering
    \includegraphics[width=0.95\linewidth]{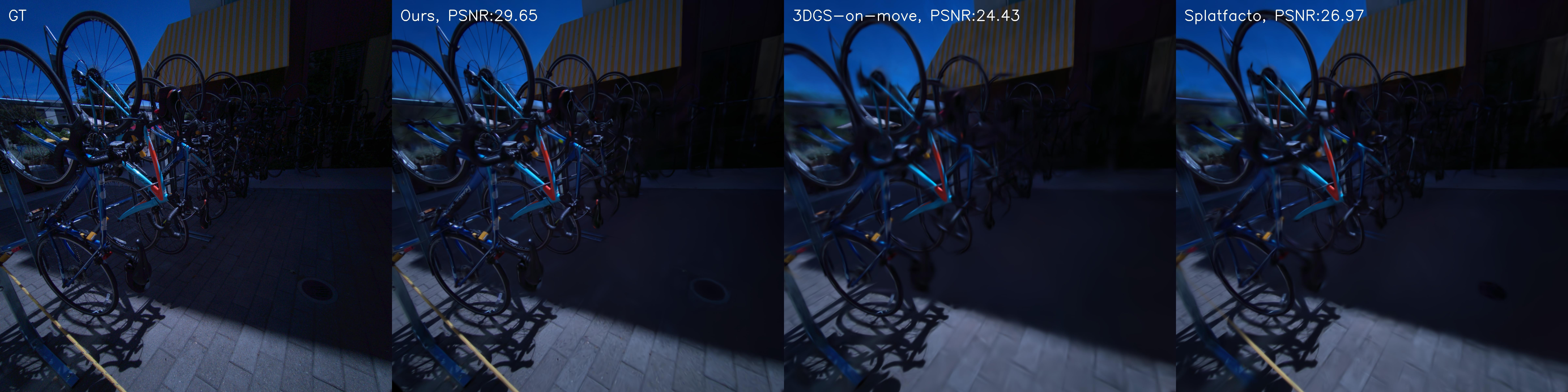}
    \includegraphics[width=0.95\linewidth]{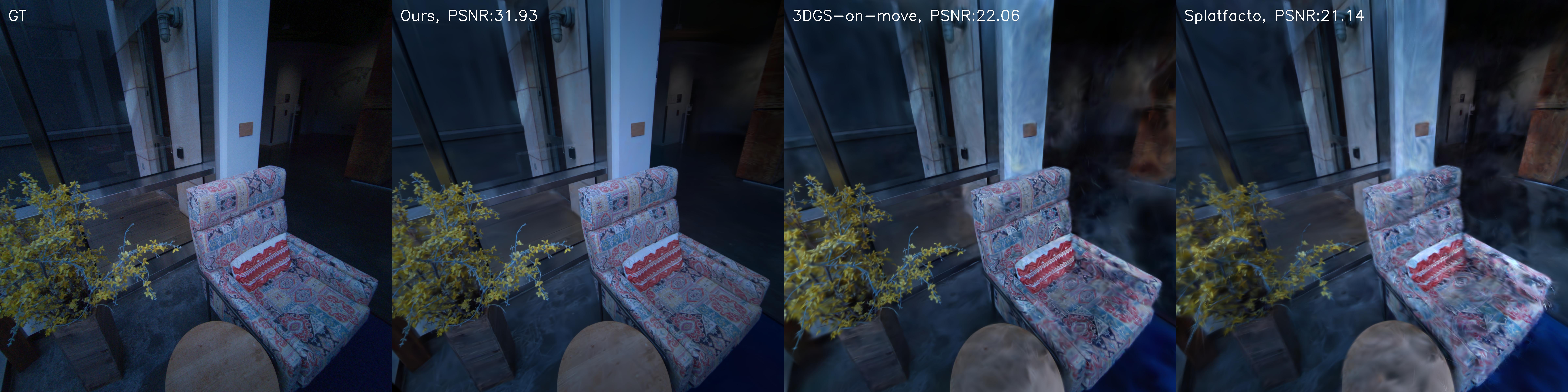}
    \includegraphics[width=0.95\linewidth]{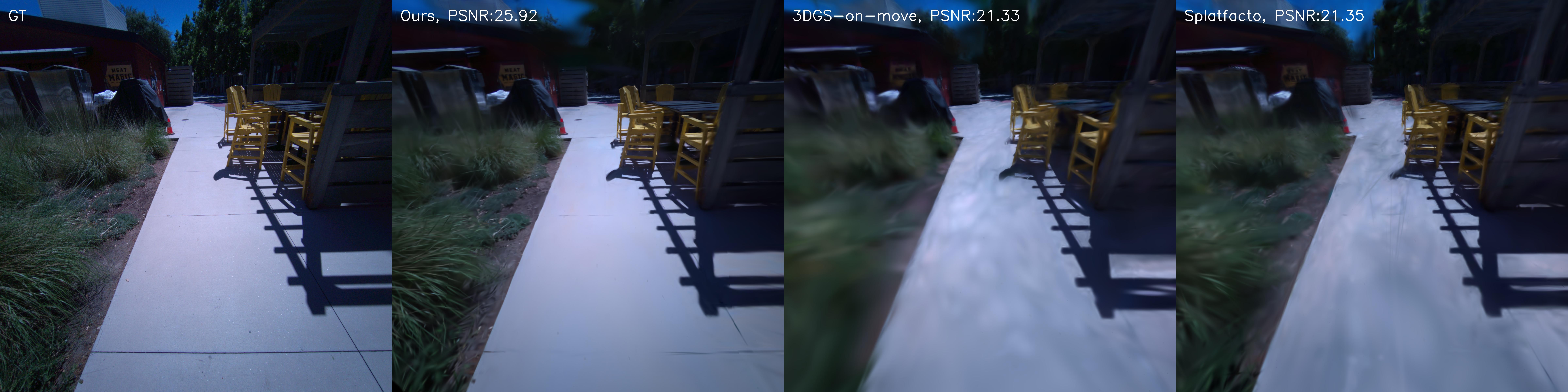}
    \includegraphics[width=0.95\linewidth]{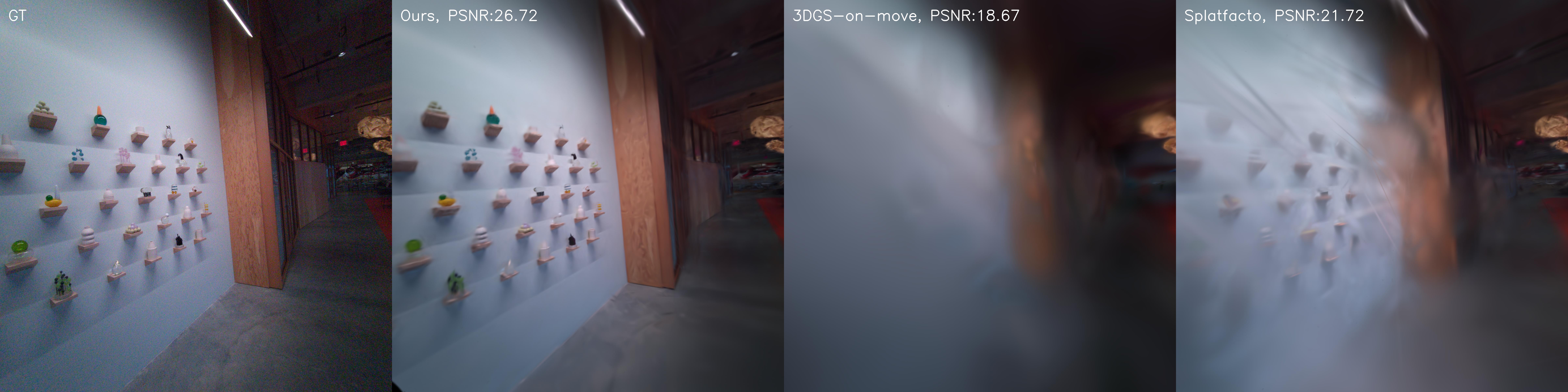}
    \includegraphics[width=0.95\linewidth]{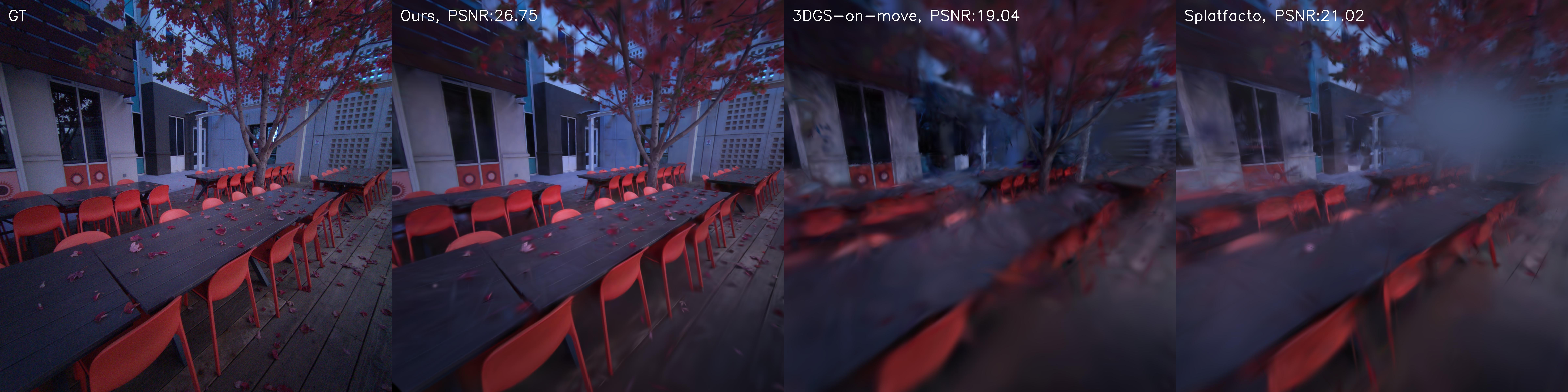}
    \vspace{-1em}
    \caption{Qualitative comparisons to baseline approaches Splatfacto and 3DGS-on-move.}
    \label{fig:qualitative_comparisons_baselines}
\end{figure*}
\begin{figure*}
    \centering
    \includegraphics[width=0.95\linewidth]{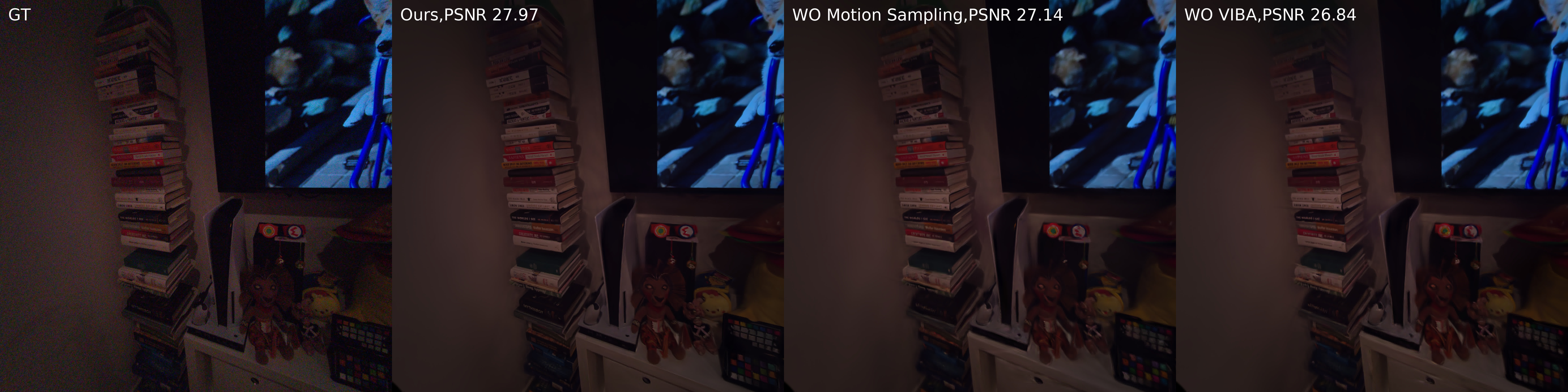}
    \includegraphics[width=0.95\linewidth]{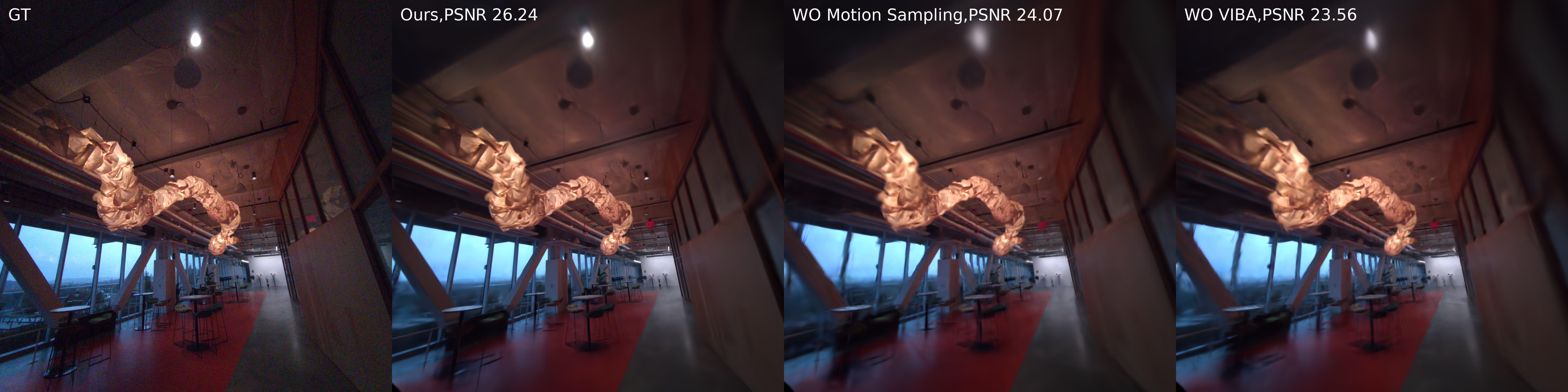}
    \includegraphics[width=0.95\linewidth]{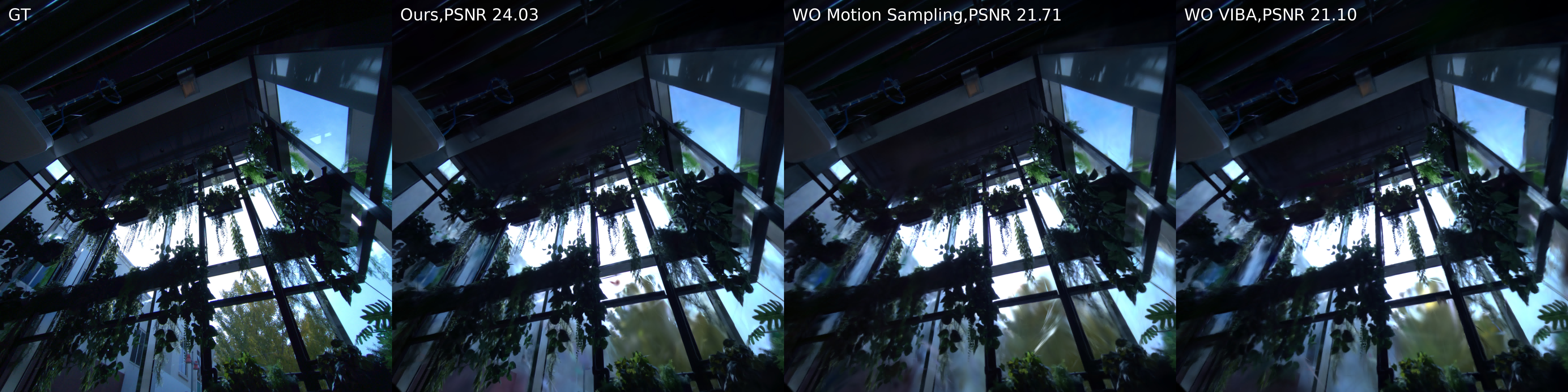}
    \includegraphics[width=0.95\linewidth]{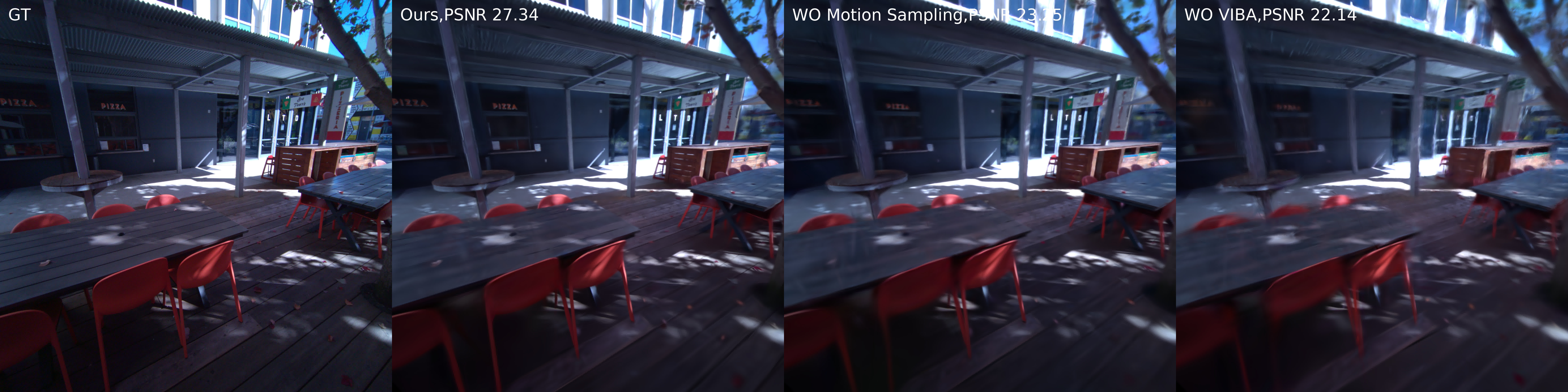}
    \includegraphics[width=0.95\linewidth]{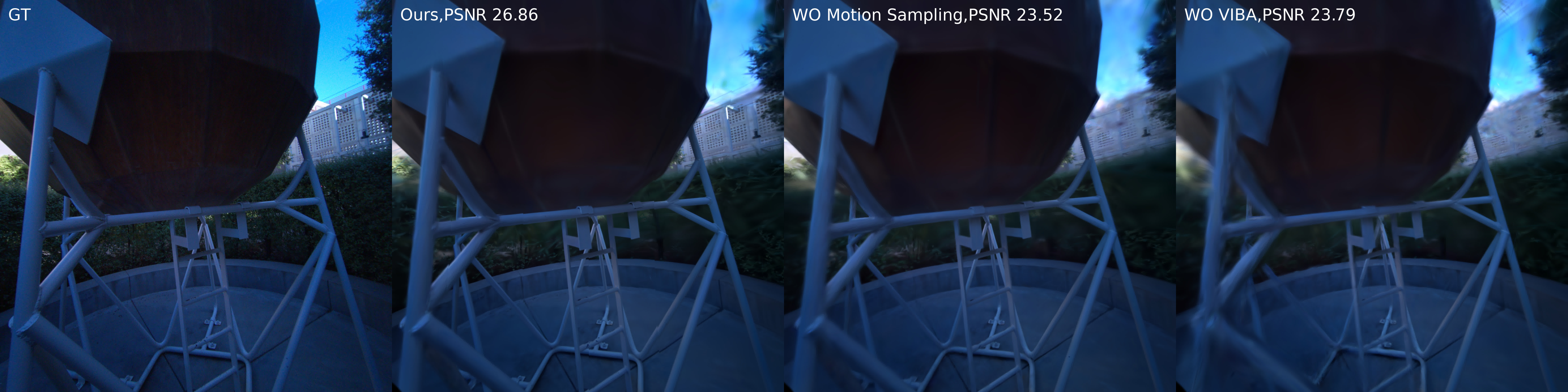}
    \vspace{-1em}
    \caption{Qualitative comparisons of ablations. Better visualized in full resolution.}
    \label{fig:qualitative_comparisions_ablation}
\end{figure*}

\paragraph{Results.} Table~\ref{tab:aria_scene} demonstrates the quantitative evaluations on Aria scene dataset, in both outdoor and indoor recordings, which includes the comparison to the baselines and ablations. Table~\ref{tab:dtc_evaluation} provides the comparison on the DTC dataset. Fig.~\ref{fig:qualitative_comparisons_baselines} show a qualitative comparisons of our method to baselines. Fig.~\ref{fig:qualitative_comparisions_ablation} compares our ablation settings. Fig.~\ref{fig:teaser} highlights the high dynamic range noise-free reconstruction in indoor environment using fast exposure captures. We summarize the findings in the following.

\paragraph{Comparison to existing work.} Compared to both splatfacto and 3DGS-on-move, our method significantly outperform them in both quantitative and qualitative comparisons. Both baselines fail to recover scene details and create significant floaters in the scene. Our solution without using VIBA or camera motion sampling also outperforms them in most scenes. Although 3DGS-on-move rasterizes a physical image model considering both rolling-shutter and blur, its joint optimization with Gaussians splatting does not necessarily provide better visual quality compared to the splatfacto. Both methods come close in a small dense captured scene (Livingroom), but we observe larger performance gap for scenes with increasing scale and complexity.

\paragraph{The effect of VIBA and motion sampling.} We observe a big improvement in visual quality (2-3db in PSNR) when using optimized calibration from VIBA, and we can also observe consistent improvement (1db in PSNR) when using the motion sampling to compensate the camera rolling-shutter effect. In DTC evaluation, we can observe using VIBA and motion sampling in color will also contribute to better geometry modeling. 

\paragraph{The effect of modeling gamma compressed scene radiance.} Explicitly modeling the radiance in gamma space can boost visual quality in all scenes and have more significantly impact in outdoor scenes. As mentioned in Section~\ref{sec:method}, it helps to model scenes with higher dynamic range. 

\paragraph{Generalization to other headset.} We apply the same process to one recording using Meta Quest 3 dataset and report the quantitative comparison in Table~\ref{tab:quest_scene} and perform the ablation study on 3D-GS baselines. Similar as the observation in Aria datasets, we can consistently observe the performance improvement measured in all metrics when using VIBA and motion sampling.

\paragraph{High dynamic range rendering.} Our reconstruction in low-light indoor scenario preserve the dynamic range of the scene, which can produce enhanced rendering after reconstruction. In Fig.~\ref{fig:teaser}, we show a visual example to simulate a rendering camera with 3x gain. We can see the improved clarity of details and text despite the input video is noisy and dark.


\begin{table}[t]
\centering
\caption{Quantative evaluations on Meta Quest 3 recording dataset.}
\begin{tabular}{lccc}
\toprule
      & PSNR $\uparrow$ & SSIM $\uparrow$ & LPIPS $\downarrow$  \\ \midrule
Ours (3D-GS) & \textbf{29.54} & \textbf{0.9147} & \textbf{0.2271} \\
w/o VIBA & 27.27 & 0.8737 & 0.2731 \\
w/o motion sampling & 28.85 & 0.9012 & 0.2344 \\
\bottomrule
\end{tabular}

\label{tab:quest_scene}
\end{table}



\begin{figure*}
    \centering
    \begin{subfigure}{0.33\linewidth}
        \centering
        \includegraphics[width=0.95\linewidth]{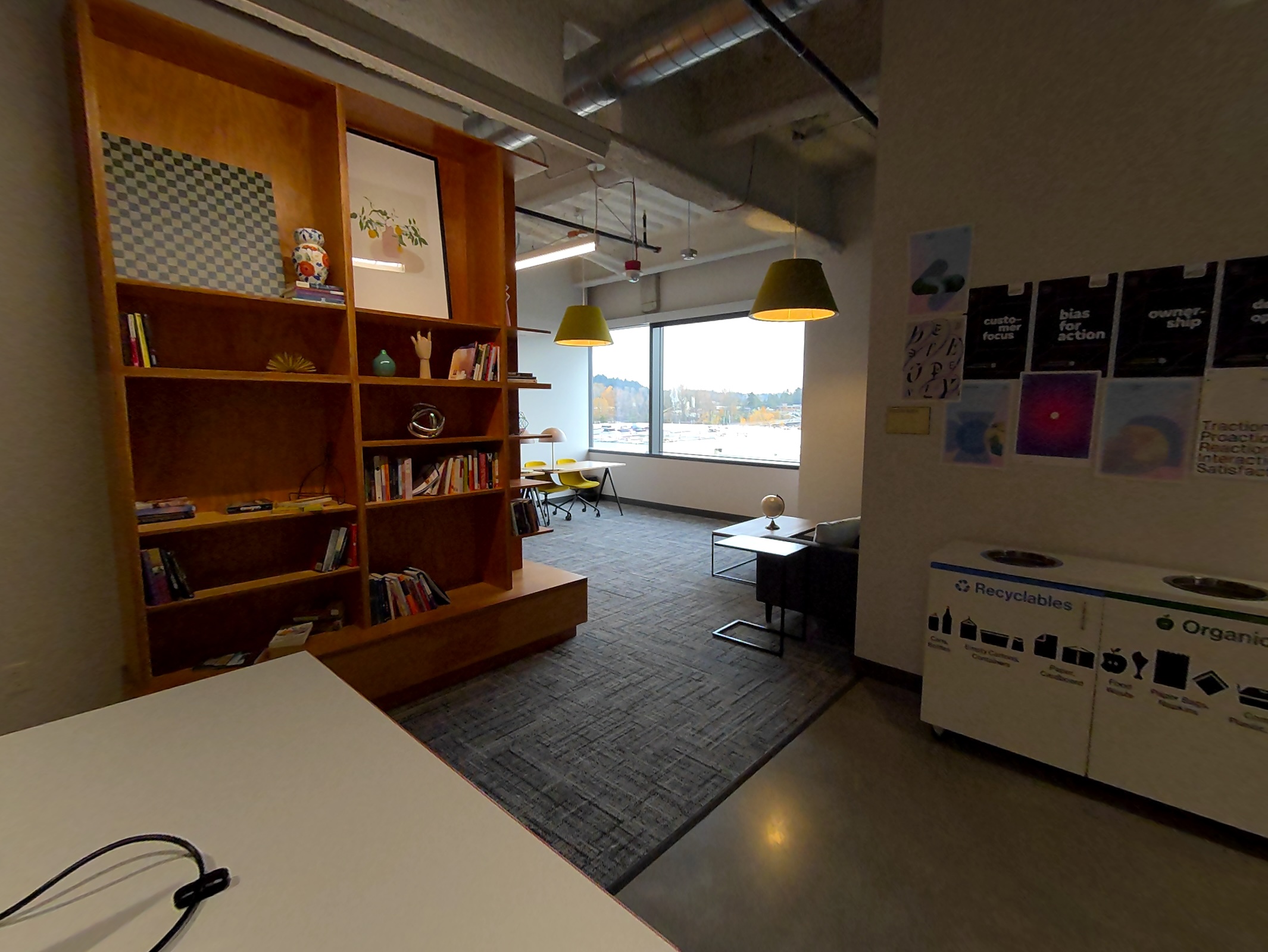}
        \caption{Ground Truth}
    \end{subfigure}
    \hfill
    \begin{subfigure}{0.33\linewidth}
        \centering
        \includegraphics[width=0.95\linewidth]{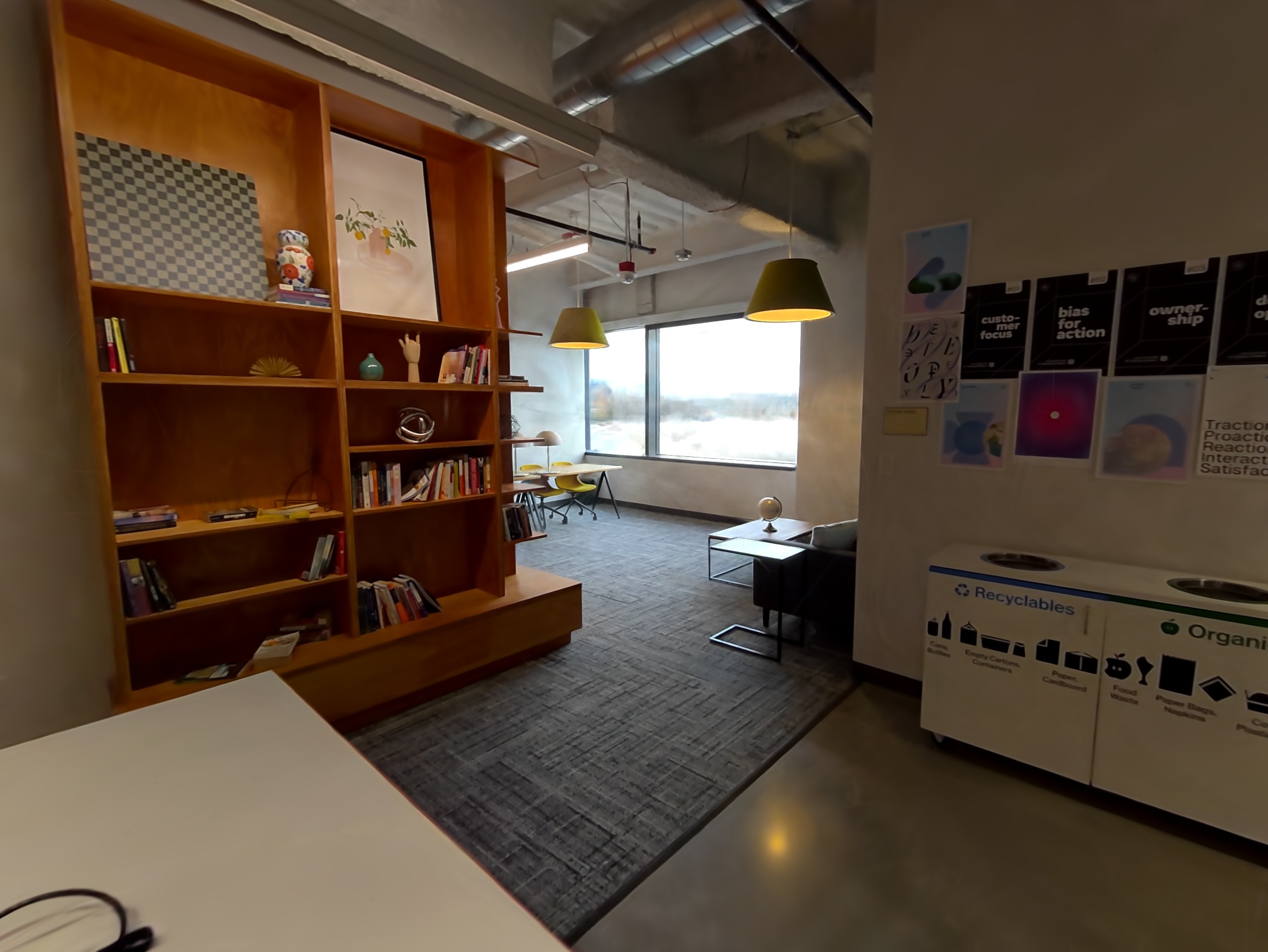}
        \caption{Ours}
    \end{subfigure}
    \hfill
    \begin{subfigure}{0.33\linewidth}
        \centering
        \includegraphics[width=0.95\linewidth]{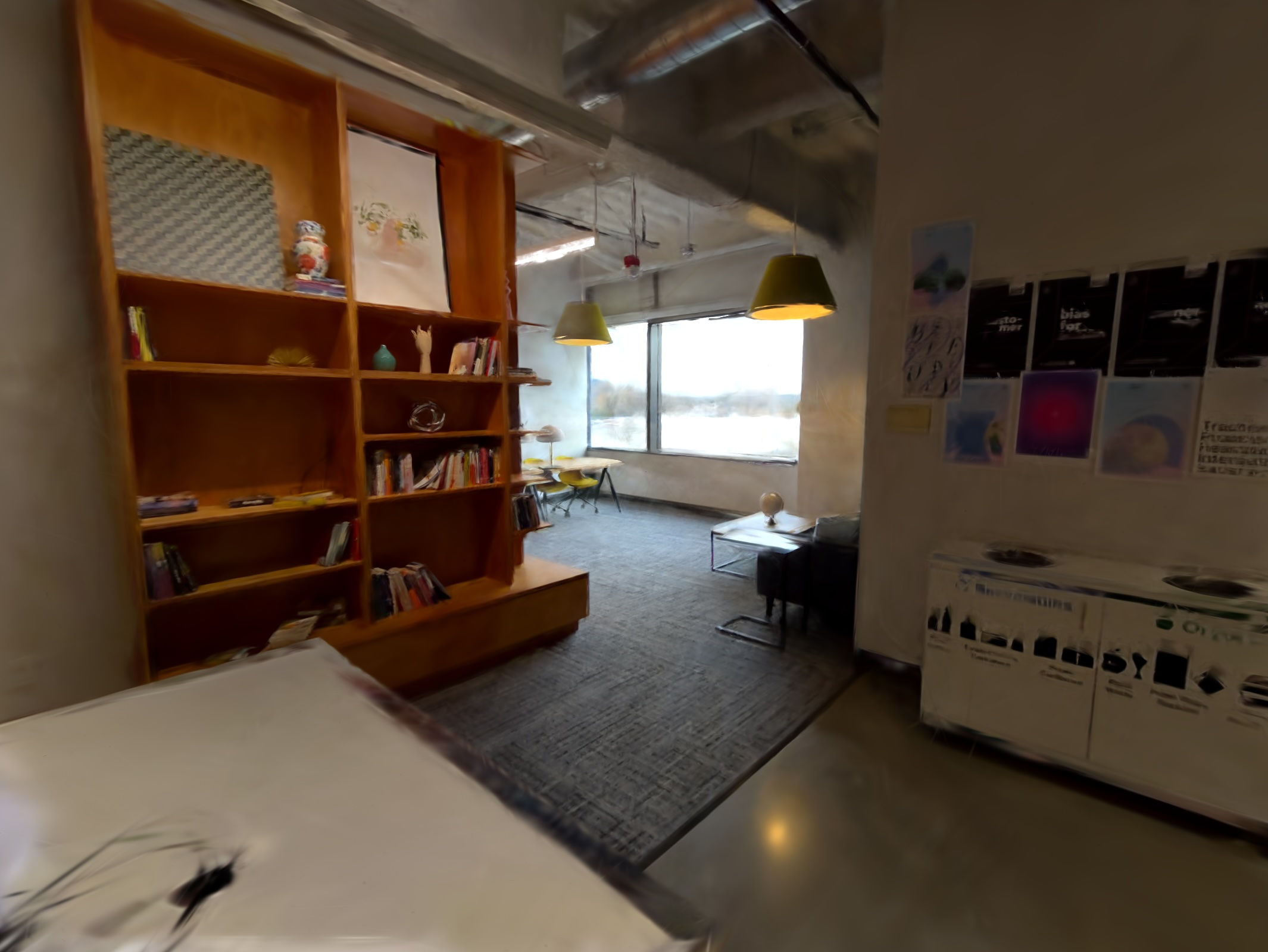}
        \caption{Without VIBA}
    \end{subfigure}
    \caption{Qualitative comparison on a Quest 3 device. We can reconstruct scenes with sharper details using VIBA.}
    \label{fig:qualitative_3dgs_2dgs}
\end{figure*}




\section{Conclusion}
\label{sec:conclusion}

In this paper, we describe a system for capturing videos using an egocentric device and a method for reconstructing photorealistic scenes. We argue the importance of correctly calibrating and modeling the physical image formation model from first principles. In our evaluation, our results produce high quality rendering with sharp details while existing methods fail to do so.
The proposed solution is built on an open-source egocentric glasses platform, and we tested it across scenes with varying complexity and lighting conditions. We further validated our approach on a commercial headset, leading to consistent conclusions across different platforms. With the rising demand for lightweight egocentric glasses, we believe our method can benefit various applications and inspire future methods to design better scene reconstruction algorithm and hardware end-to-end.

\paragraph{Limitations and future work.} Firstly, while our method surpasses current state-of-the-art techniques in scene reconstruction, reconstructing any scene from any human trajectory remains an unsolved challenge. Unlike static captures, egocentric video may lack sufficient view coverage, leading to reconstruction artifacts. Recent methods using sparse views show promise and could inform future improvements. Secondly, like most reconstruction algorithms, we assume a static scene, which is a significant limitation. Human body parts, shadows, illumination changes, and scene motions are often present and should be addressed in future work to enhance scalability. Lastly, our system struggles in extremely low light conditions (<50 lux), resulting in insufficient signal capture and failure. Future advancements in image sensing and reconstruction algorithms may better address this issue.

\begin{acks}
We would like to thank the team behind Project Aria, machine perception services and open source team. They provided the open source egocentric device platform and made this research possible. We thank David Caruso for helping inspecting tools using visual inertial bundle adjustment, Sam Zhou, Rajvi Shah for their help on Quest3 data experiments. We also thank the anonymous reviewers for their insightful feedback.     
\end{acks}


\bibliographystyle{ACM-Reference-Format}




\clearpage

\appendix

\section{Details of the Aria scene dataset}

\begin{figure*}[t]
    \centering
    \includegraphics[width=0.98\textwidth]{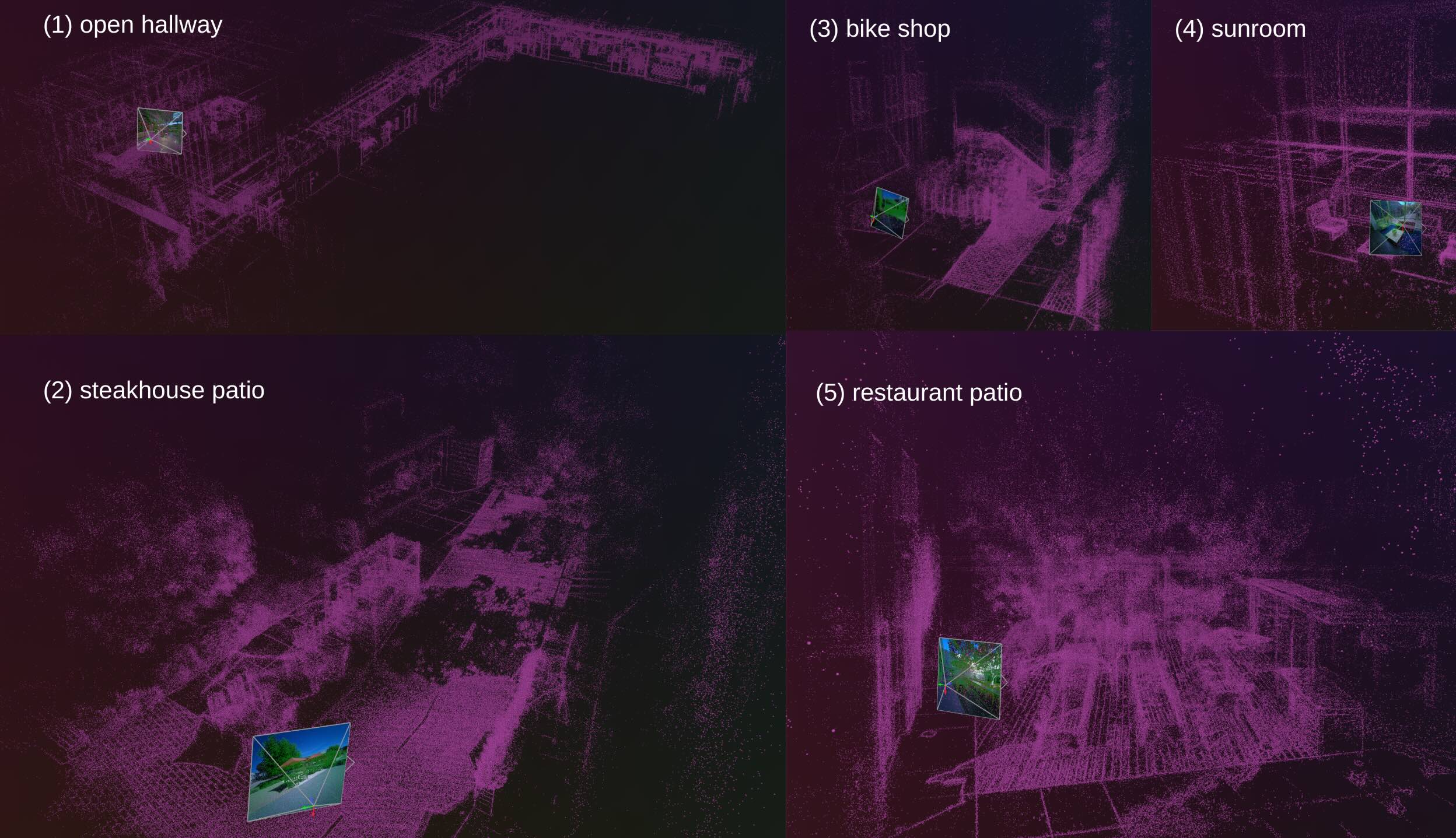}
    \caption{The visualized point cloud using semi-dense point cloud and posed RGB view from a few scenes. We cover scenes within a large indoor building as (1) open hallway, large open spaces in outdoor as (2) steakhouse patio, outdoor with complex thin structures as (3) bike shop, indoor scene with big transparent window as (4) sunroom, and large outdoor space with complex shading as (5) restaurant patio.}
    \label{fig:mps_example}
\end{figure*}

We collect the egocentric data under different lighting conditions, scene environment and with different type of device motions.
Table \ref{tab:aria_dataset_statistics} provides a summary of the collected dataset statistics. 
In general, each recording contains more number of frames compared to common used existing dataset \cite{barron2022mipnerf360,barron2023zipnerf}. We do not perform additional filtering to remove frames within the video. Fig.\ref{fig:mps_example} shows the structure of a few exemplar scene using the point cloud. This also features one challenge of egocentric recording that differentiates from static multi-view image captures. Human are constantly in motion in an open real-world environment. As our experiment results show, it brings challenges to existing well-established baselines. 

\begin{table}[t]
\centering
\caption{Aria scene dataset statistics. }
\begin{tabular}{lll}
\toprule
      & \#Frames & scenario \\ \midrule
bike shop & 1611 &  outdoor, sunny day \\
steakhouse patio & 1725 & outdoor, sunny day \\
pop-up shop & 1189 & outdoor, sunny day \\
sunroom & 691 & indoor with transparent glasses \\
garden & 1501 & outdoor, sunny day \\
restaurant patio & 1399 & outdoor, under shades \\
library & 2817 & indoor, dim light, 200- lux \\
plant hallway & 1404  & indoor, with windows, 500- lux \\
open hallway & 2730 & indoor, dim light, 200- lux \\
micro Kitchen & 2952 & indoor, dim light, 200- lux \\
multi-floor & 2193 & indoor, dim light, 200- lux \\
livingroom & 1517 & indoor, dim light, 200- lux \\
\bottomrule
\end{tabular}
\label{tab:aria_dataset_statistics}
\end{table}

\section{Implementation Details}

\paragraph{Algorithm input} To summarize the preprocessed data we used in our reconstruction model, our algorithm utilizes the following information:

\begin{itemize}
    \item Rectified RGB images.
    \item 6DoF high-frequency (1kHz-rate) device trajectory and RGB sensor calibration obtained from the location tool provided by the Project Aria machine perception service, using VIBA or not. The trajectory is denoted as $f_{T}(t)$.
    \item Per-frame image gain, exposure value, and image read-out start and end timestamps, extracted from the image metadata.
    \item A rectified RGB lens shading image.
    \item A rectified per-pixel index ratio image determining the exact timing of the pixel, used for rolling-shutter lookup as referenced in Equation 4 in the main paper.
    \item A semi-dense point cloud from the Project Aria tool to initialize the 3D Gaussians.
\end{itemize}

We employ the algorithm described in Section 5 of the main paper to reconstruct the 3D Gaussians using the information outlined above.

\begin{figure}
    \centering
    \begin{subfigure}{0.45\linewidth}
        \centering
        \includegraphics[width=0.9\linewidth]{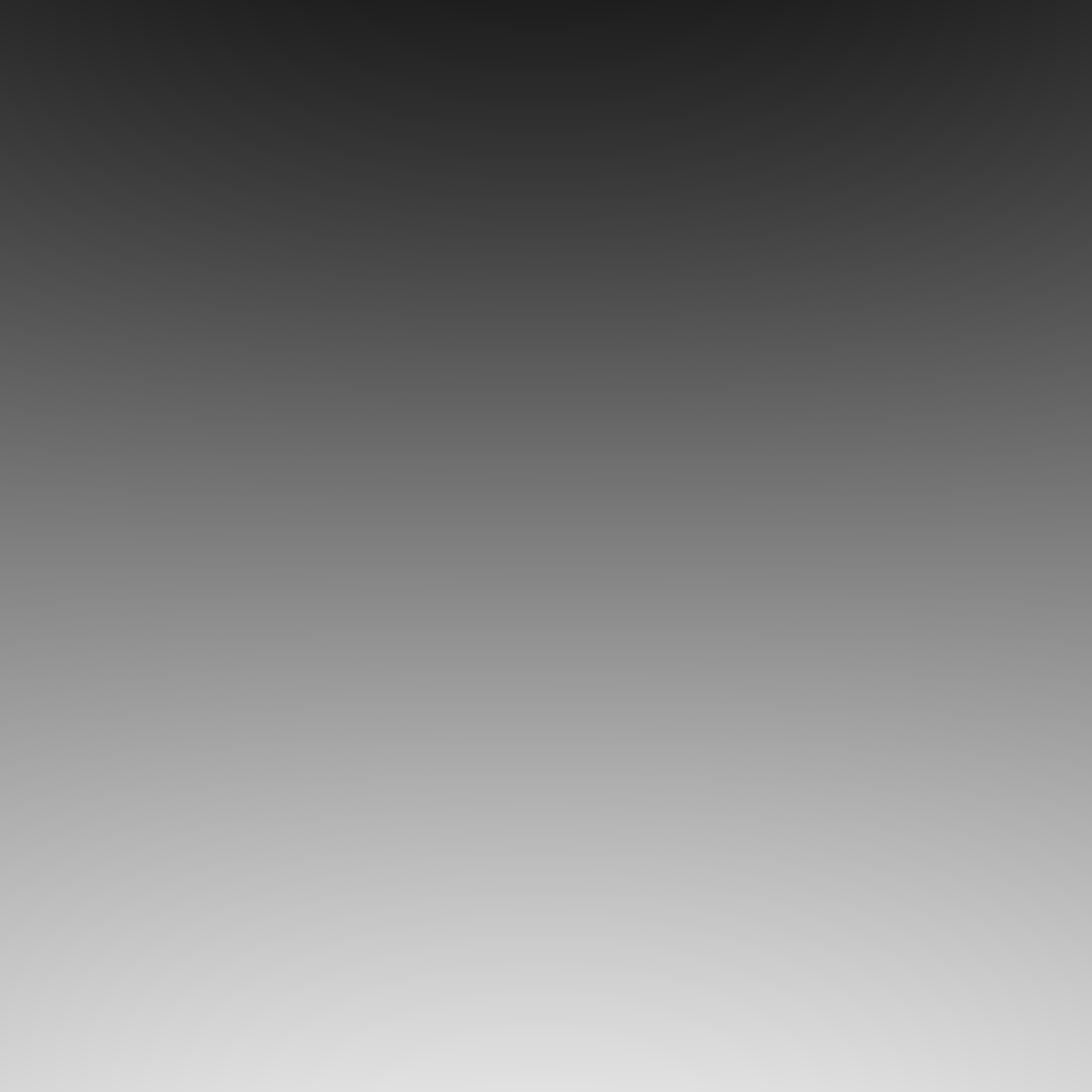}
        \caption{Linearly rectified}
    \end{subfigure}
    \hfill
    \begin{subfigure}{0.45\linewidth}
        \centering
        \includegraphics[width=0.9\linewidth]{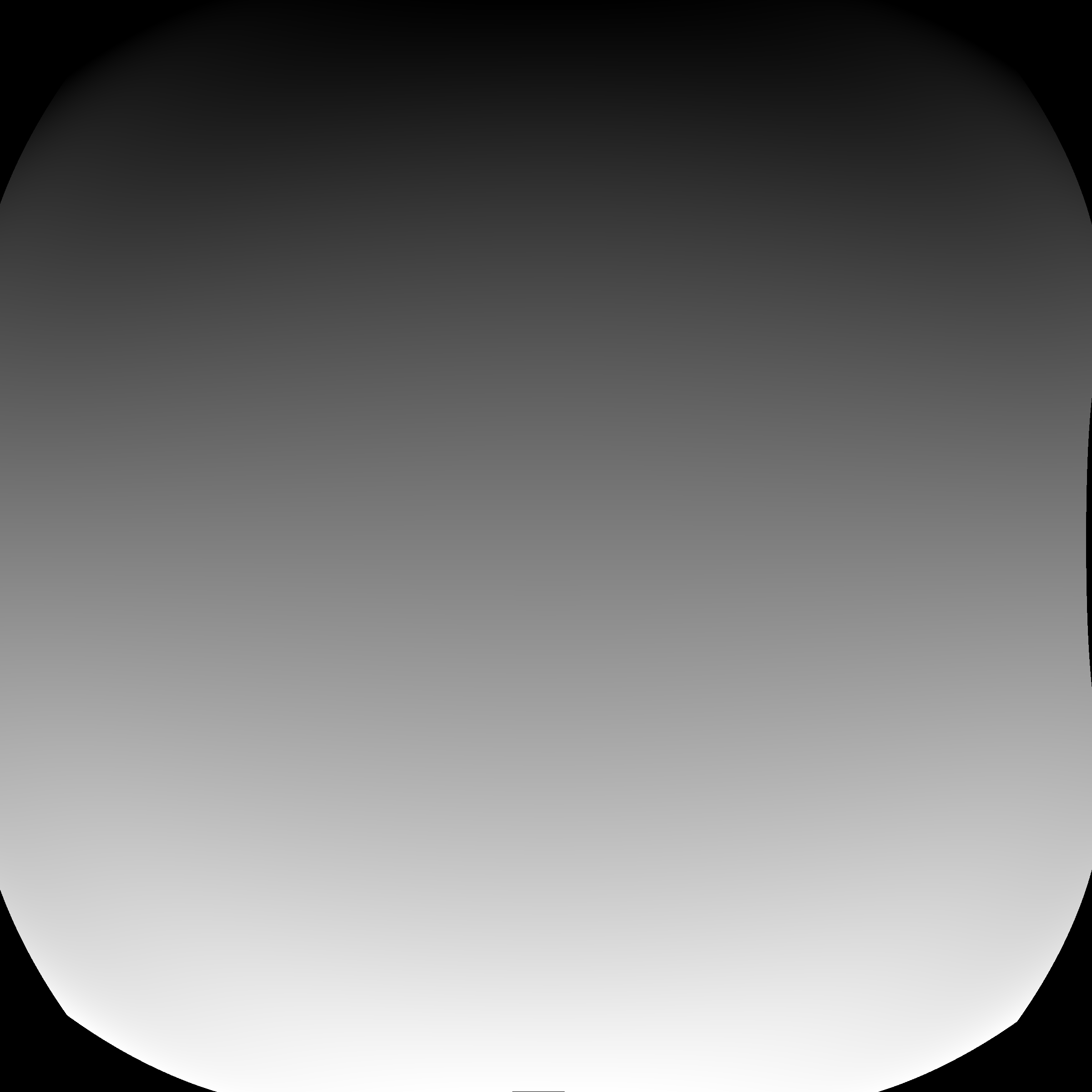}
        \caption{Fisheye rectified}
    \end{subfigure}
    \caption{An example visualization of the image index image being rectified using a linear camera model (a) and fisheye camera model (b). We represent it as a monochrome image with 1 represent the first row, and 255 represent the last row in original image. When in black (0), it means the pixels are out of origional observation, and we mask them out during training.}
    \label{fig:motion_index_image}
\end{figure}

\paragraph{Preprocessing Steps} To generate the algorithm input data, we process all the recorded datasets in the following order.
\begin{enumerate}
    \item We collect the Project Aria data in the vrs format. Then we run the machine perception service tool provided by Project Aria platform. VIBA is handled as one option flag at this step. For the ablation study that do not use VIBA, we turn off the flag. The output remains the same for both options. We acquire the high-frequency device trajectory, high frequency online calibration and semi-dense point cloud. The high frequency online calibration contains the RGB camera intrinsics and extrinsic at IMU rate only when VIBA is used. Otherwise, we can only use device calibration to estimate the RGB camera calibration. 
    \item After acquire all the device information and calibration, we featch the RGB camera timestamp its metadata in exposure, device timestamp, sensor readout time, and calculate its derived calibration in intrinsics and extrinsics. Note for rolling-shutter camera, we do not represent RGB extrinsic using a single camera pose. We only calculate the timestamps information for all the rows with their exposure values, and then calculate the pixel pose using the continuous trajectory on the fly. We also pre-calculate the pose for the center row of the image, which is used as the pose input when rolling-shutter is not considered.
    \item Then we rectify the raw images using a chosen conventional camera model that the rasterization algorithm will support. We consider the pinhole camera model or the equidistant fisheye model \cite{liao2024fisheyegs}, which are both supported in GSplats\cite{ye2024gsplats}. Through testing, we do not observe significant visual difference choosing between pinhole or fisheye camera model. We use pinhole as the convention for all the studies. The focal length and FOV is chosen based on the trade-off to preserve maximum number of pixels. We use 1200 as the focal value and rectify the input 2880x2880 images into 2400x2400. All the training and evaluations are performed using these rectified images. 
    \item The rectification step will change the pixel ordering. For all information that require a pixel aligned value, we need to perform rectification at this step as well. This include rectify the lens vignette, and the proposed motion sample image. Fig.\ref{fig:motion_index_image} visualized the example of the rectified image index in both linear and fisheye mode.
    \item After rectification, we project the scene point cloud to each image and acquire the sparse depth. The point cloud from Project Aria device are calculated from the global shutter SLAM camera. They provide the static 3D anchors that do not affected by the RGB camera model. Given the sensor timestamp of RGB camera, we fetch the visible point cloud calculated from the SLAM camera views in time, and project them to the RGB image, which form the sparse depth. We use this information to calculate the reprojection error of pixels when the camera moves in time. 
\end{enumerate}

\paragraph{Training} We implement the method in pytorch. We use the rasterization kernel in GSplats and use the same training loss as vanilla 3D-GS. When render the rolling-shutter image, we render it as a batched rendering and gather the final image regarding the image index to the batch index. A batch process will consume larger memory, which can also be replaced by an iteration when GPU memory is constrained. We trained all the model at 2400x2400 resolution using a single GPU in A6000 or A100. To speed up training, we perform the rolling-shutter compensation after 7.5K iterations (total 30K). 


\section{Additional Results}

\begin{figure*}
    \centering
    \includegraphics[width=0.45\linewidth]{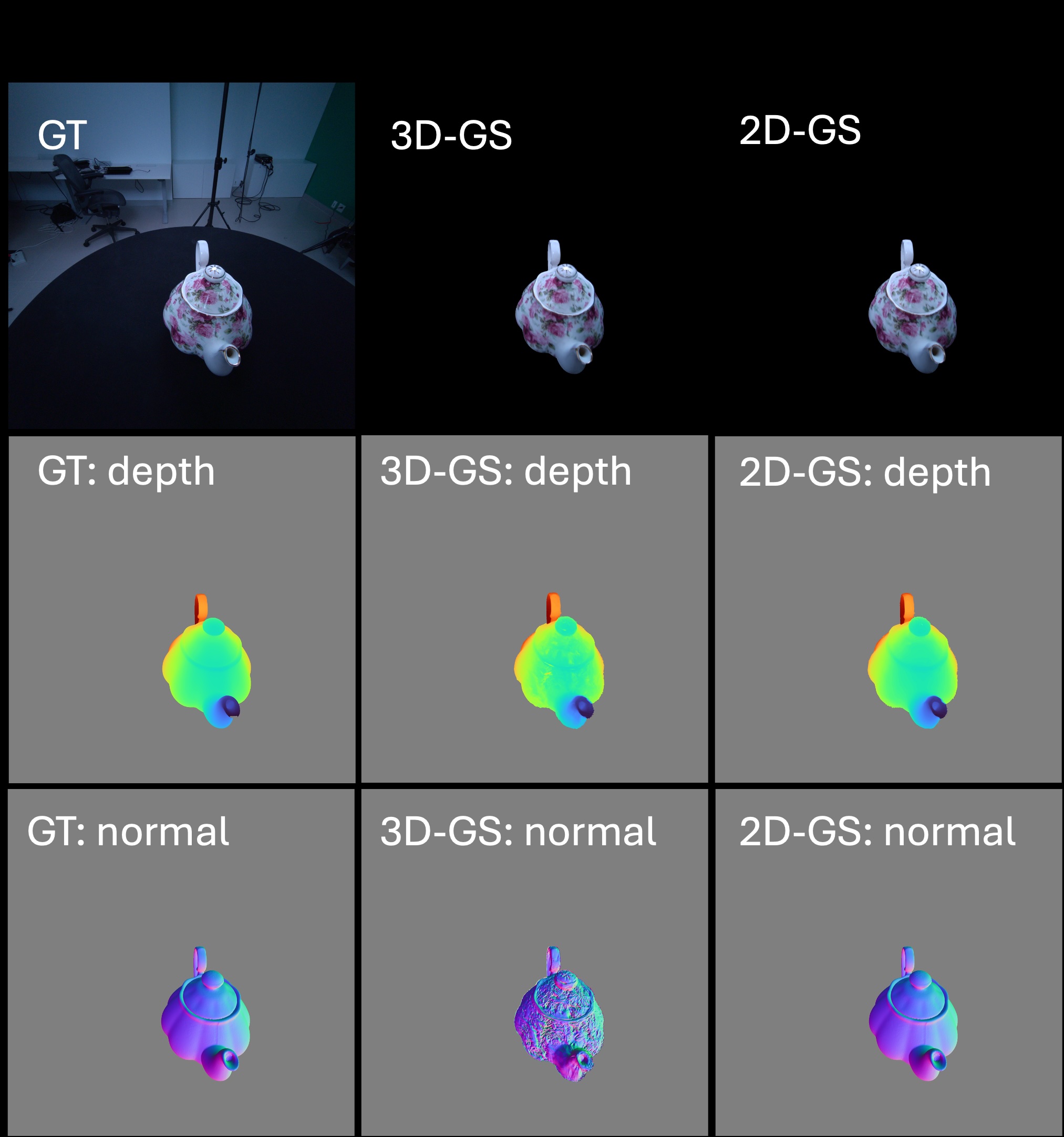}
    \includegraphics[width=0.45\linewidth]{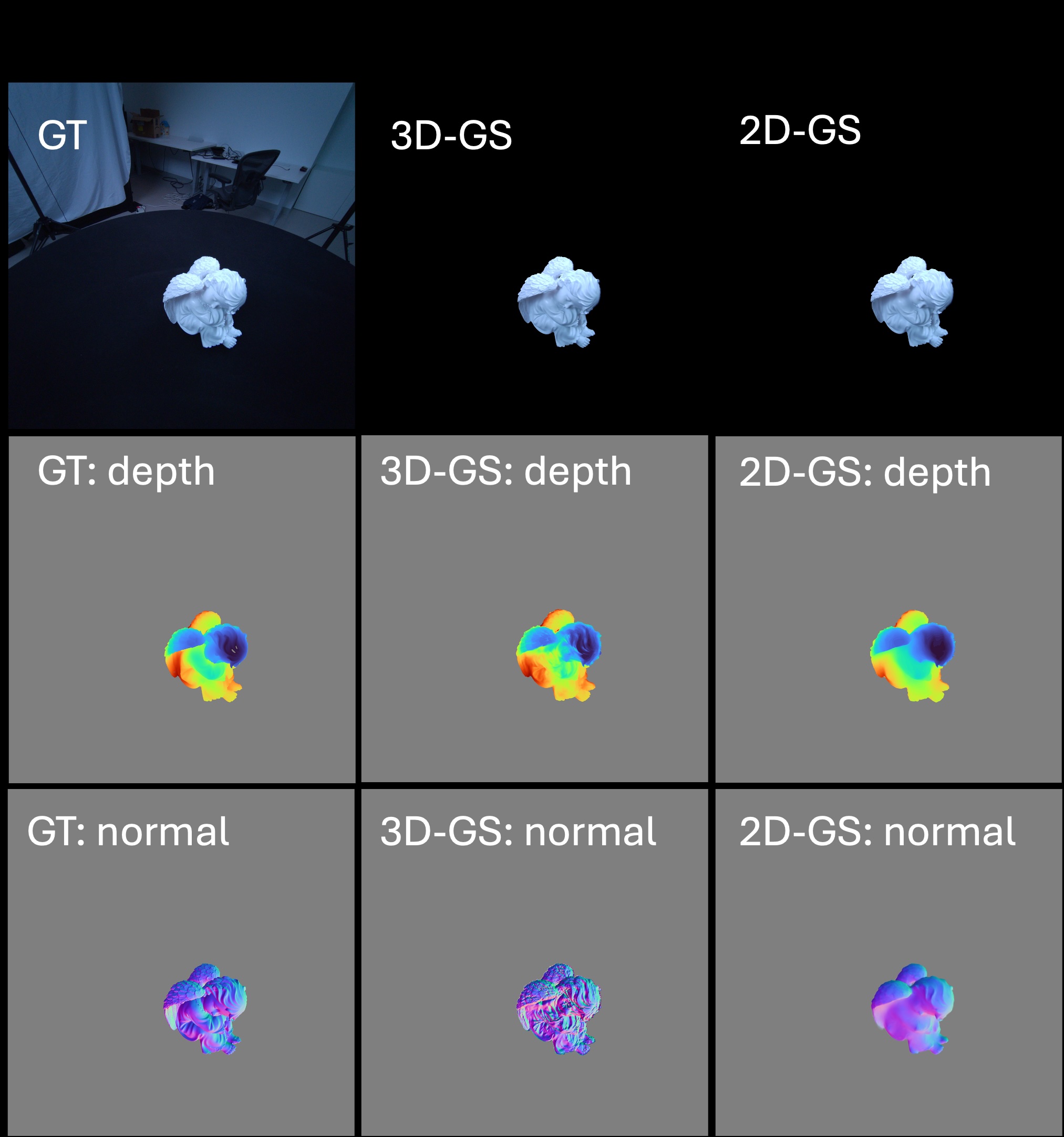}
    \includegraphics[width=0.45\linewidth]{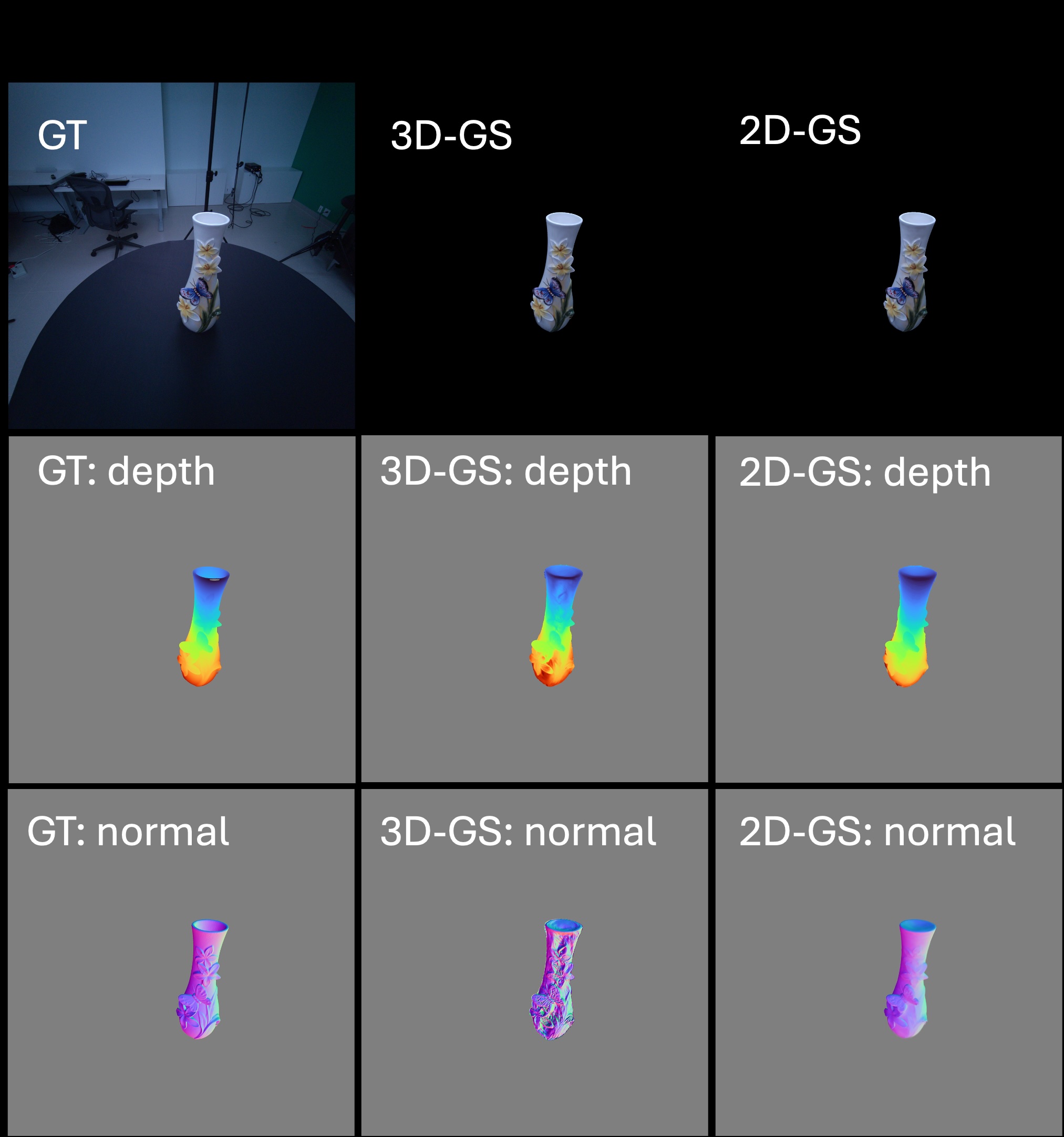}
    \includegraphics[width=0.45\linewidth]{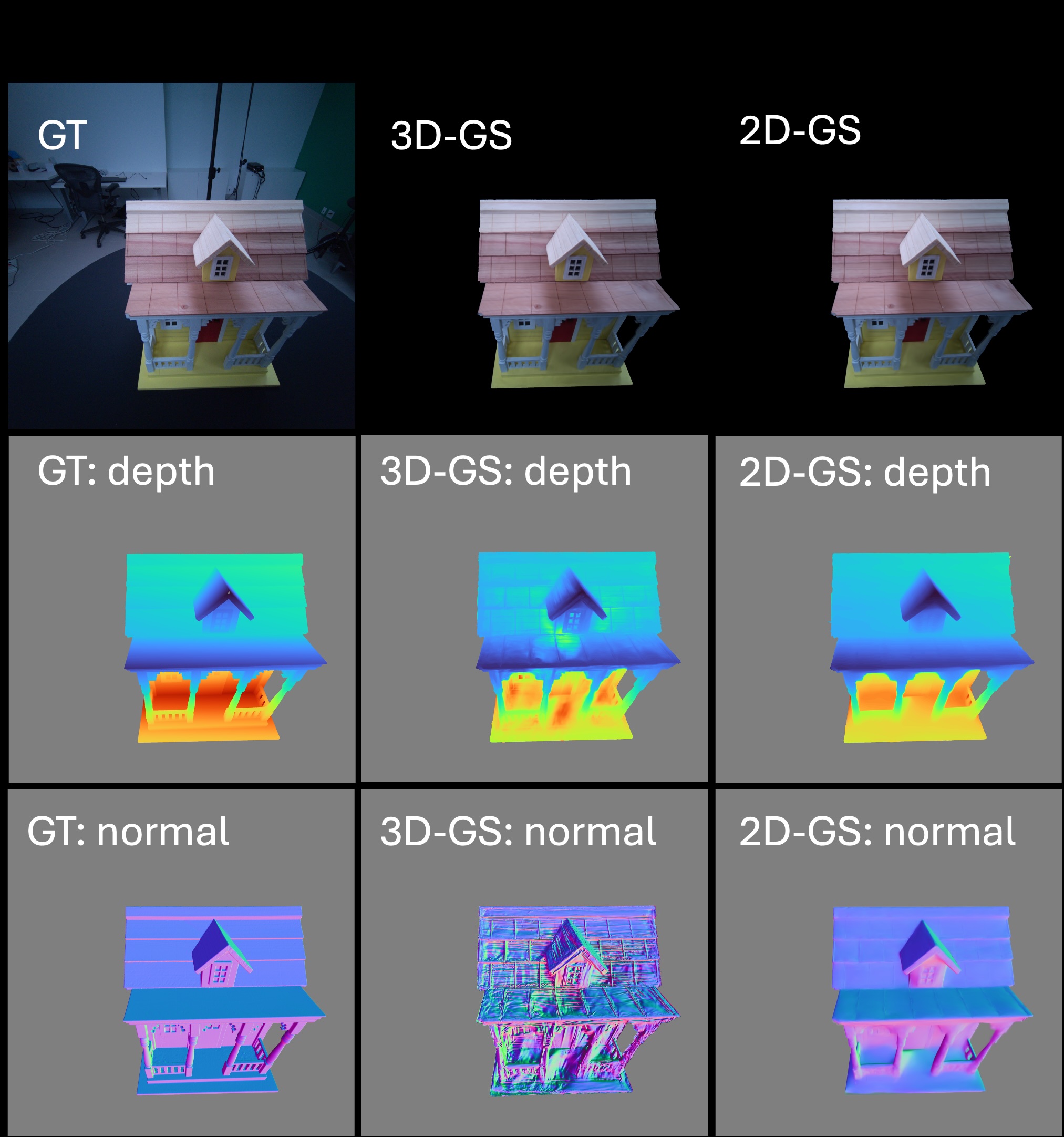}
    \caption{Visualize of the appearance and geometry reconstruction of our method using 3D-GS and 2D-GS\cite{huang20242dgs}. The ground truth is acquired using the modalities of rendered images, depth and normal. For small object, using 2D-GS can further enhance geometry reconstruction.}
    \label{fig:aria_baseline_comparisons}
\end{figure*}

\textbf{DTC dataset visualization.} In Fig.\ref{fig:aria_baseline_comparisons}, we provide a visualization of our full rolling-shutter aware model in DTC dataset using both 3D Guassian rasterization and 2D Gaussian rasterization. As Table 3 and Fig.\ref{fig:aria_baseline_comparisons} in the main paper indicate, using 2D-GS can dramatically improve the geometry reconstruction (as seen in depth and normal). We can incorporate this variant of 2D-GS in a different application by simply replacing the rasterization kernel, while existing work that required specific Gaussian parameterization \cite{seiskari2024gaussian_on_the_move} can not. 

\textbf{Video} Refer to our video asset on \href{https://www.projectaria.com/photoreal-reconstruction/}{Project Aria Photoreal Reconstruction}.

\end{document}